%% file: iclr2027_conference.tex
\documentclass{article} 
\usepackage{iclr2027_conference,times}

\input{math_commands.tex}

\usepackage{hyperref}
\usepackage{url}
\usepackage{booktabs}
\definecolor{LedgerBlue}{RGB}{235,243,252}
\usepackage{amsmath}
\usepackage{booktabs}
\usepackage{graphicx}
\usepackage{tabularx}
\usepackage{array}
\usepackage{wrapfig}
\usepackage[table]{xcolor}
\usepackage{pifont}
\usepackage{threeparttable}
\definecolor{ACTBlue}{HTML}{EDF4FC}
\definecolor{ACTGain}{HTML}{147D64}
\definecolor{ACTLoss}{HTML}{AB5147}

\usepackage{tabularx}
\usepackage{threeparttable}
\usepackage[table]{xcolor}

\usepackage{wrapfig}
\usepackage{booktabs}
\usepackage{capt-of}
\providecolor{MethodBlue}{HTML}{1F4E79}
\providecommand{\methodinnovation}[2]{%
\noindent\textcolor{MethodBlue}{\textbf{\textit{[#1]}}}\ \textbf{#2}}

\definecolor{FDHeader}{HTML}{EAF0F7}
\definecolor{FDStripe}{HTML}{F7F9FC}

\providecolor{ResultBlue}{HTML}{1F4E79}

\newcolumntype{Y}{>{\centering\arraybackslash}p{1.12cm}}

\definecolor{YesGreen}{HTML}{147D64}
\definecolor{NoRed}{HTML}{B23A48}
\definecolor{OursBlue}{HTML}{EAF3FF}
\providecolor{MethodBlue}{HTML}{1F4E79}

\newcommand{\cmark}{\textcolor{YesGreen}{\ding{51}}}
\newcommand{\xmark}{\textcolor{NoRed}{\ding{55}}}

\hypersetup{
  colorlinks=true,
  linkcolor=red,
  citecolor=cyan,
  filecolor=magenta,
  urlcolor=magenta,
}

\title{DuplexLedger: Benchmarking Transactional Memory in Full-Duplex Speech Agents}

\author{%
  \parbox{0.95\textwidth}{%
    \centering
    Yuxuan Li$^{1,*}$,
    Xiangchen Guan$^{2,*}$,
    Kunrong Li$^{3}$,
    Jin Wang$^{4}$,
    Kaiwen Luo$^{1}$,
    Yang Xiao$^{5}$,
    Yufei Guo$^{6}$,
    Zhisheng Chen$^{1}$,
    Liang Lin$^{1,\dagger}$,
    Kun Wang$^{1,\dagger}$\\[6pt]
    {\normalfont\small
      $^{1}$Nanyang Technological University
      $^{2}$Peking University
      $^{3}$Beijing University of Technology
      $^{4}$Xidian University
      $^{5}$University of Melbourne
      $^{6}$China Aerospace Science and Industry Corporation
      \\[5pt]
      $^{*}$Equal contribution.
      \quad
      $^{\dagger}$Corresponding authors.
    }
  }
}

\iclrfinalcopy
\begin{document}

\maketitle

\begin{abstract}
Full-duplex speech agents must update persistent memory while listening and speaking, yet it remains unclear which events should be stored, revised, or excluded. Existing benchmarks isolate interaction timing from memory outcomes, obscuring the state transitions behind a correct answer. We introduce \textsf{DuplexLedger}, a bilingual benchmark of 1,094 persistent sessions aligning four full-duplex scenarios with event-level \textsc{Write}/\textsc{Update}/\textsc{Ignore}/\textsc{Retain} supervision, explicit pre/post memory states, and 6,564 same-session questions and checks. Evaluating five full-duplex models reveals an interaction--memory gap: no system combines strong boundary control with coherent final-state memory, and the best final-state fact coverage is 2.01\%. We propose \textsf{ACT-Mem}, an 8.92M-parameter sidecar comprising an event controller, a versioned transactional ledger, and a typed retriever coupled with a frozen \texttt{MiniCPM}-o~4.5 backbone. On 219 held-out sessions, \textsf{ACT-Mem} increases latest-value accuracy from 4.11\% to 8.68\% while reducing stale assertions and distractor adoption. It improves interaction boundaries but slows interruption stopping. These results motivate selective, auditable full-duplex memory rather than indiscriminate transcript retention. Code is available at \url{https://github.com/lyx-researcher/DuplexLedger}.
\end{abstract}

\section{Introduction}
\label{sec:introduction}

Speech is a natural interface for agents that assist people continuously. End-to-end spoken language models now support open-ended voice interaction~\citep{zhang2023speechgpt,tang2024salmonn,chu2024qwen2audio,xie2024miniomni}, while full-duplex systems listen while speaking, enabling interruptions, backchannels, and overlapping turns~\citep{moshi,freezeomni,personaplex,lycheefd}. Yet concurrency creates a state-management problem: an utterance may introduce or revise a fact, remain incomplete, serve only as a backchannel, or be addressed to someone else. A dependable agent must decide what was said and whether it should modify persistent memory.

\begin{figure}[h!]
    \centering
    \includegraphics[width=\linewidth]{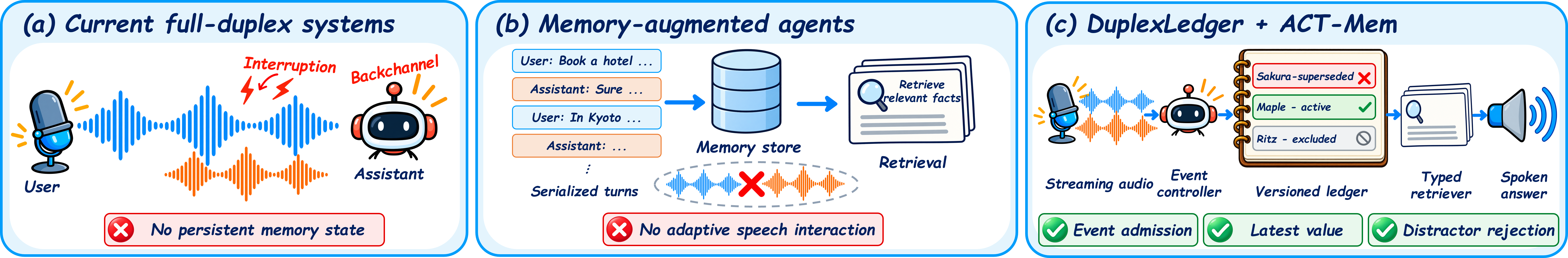}
    \caption{From isolated interaction and memory to transactional full-duplex memory.}
    \vspace{-1.0em}
    \label{fig:transactional_full-duplex_memory}
    \vspace{0.6em}
\end{figure}

Existing evaluation treats interaction and memory largely as separate capabilities. Full-duplex benchmarks emphasize turn taking, pause handling, overlap, and interruption, with recent extensions adding corrections and entity tracking~\citep{lin2025fullduplexbench,lin2026fullduplexbenchv15,wang2026humdial,lin2026fullduplexbenchv2,modi2026echochain}. Memory methods and benchmarks instead study long-horizon storage, retrieval, and updating from serialized histories~\citep{packer2023memgpt,maharana2024locomo,wu2025longmemeval,xu2025amem}, while recent speech agents begin to connect persistent memory with streaming interaction~\citep{yao2025egomem,xie2026voicemem}. The compared resources still do not jointly expose event admission, supersession, distractor exclusion, and memory-grounded spoken answers under adaptive concurrency. Without aligned state transitions, a correct answer cannot show whether the system maintained coherent memory or merely matched its current context.

We formulate this capability as \emph{transactional memory for full-duplex agents}: each event is admitted, rejected, or used to revise a versioned state. We introduce \textsf{DuplexLedger}, a bilingual benchmark of 1,094 persistent sessions aligning pause handling, speech to others, backchannels, and interruptions with \textsc{write}, \textsc{update}, \textsc{ignore}, and \textsc{retain} operations, explicit pre/post states, and same-session spoken questions. We further propose \textsf{ACT-Mem}, coupling an event controller, a versioned transactional ledger, and a typed retriever with a frozen full-duplex backbone. The controller determines admission and operation; the ledger records committed values and supersession through \textsc{prepare}, \textsc{commit}, and \textsc{abort}; and the retriever supplies active, type-appropriate evidence only for memory QA. Streaming generation remains frozen, making memory transitions explicit and auditable.

\noindent\textbf{Contributions.}
(1) We formulate full-duplex memory as selective event-level state construction.
(2) \textsf{DuplexLedger} aligns adaptive interaction, reference state transitions, and spoken QA across four scenarios, 1,094 bilingual sessions, and 6,564 questions/checks.
(3) \textsf{ACT-Mem} combines admission control, versioned memory, and typed retrieval with a frozen backbone, increasing latest-value accuracy from 4.11\% to 8.68\% while reducing stale assertions and distractor adoption.

\section{Related Work}
\label{sec:related_work}

\textbf{Full-Duplex Dialogue.}
Early spoken language models established end-to-end speech understanding and generation~\citep{zhang2023speechgpt,tang2024salmonn,chu2024qwen2audio,xie2024miniomni}. Full-duplex systems subsequently modeled synchronized channels, paired next-token prediction, or parallel audio streams to support concurrent listening and speaking~\citep{nguyen2023generative,veluri2024beyond,wang2025fullduplex,moshi}. \texttt{Freeze-Omni} retains a frozen language backbone, while \texttt{MiniCPM}-o~4.5, \texttt{PersonaPlex}, and \texttt{Lychee-FD} further target low-latency interaction, role control, and acoustic--semantic coordination~\citep{freezeomni,minicpmo45,personaplex,lycheefd}. These systems improve conversational timing, but generally provide no explicit policy for deciding which overlapping events should enter or revise persistent memory.

\textbf{Memory-Augmented Agents.}
Memory-augmented agents manage conversational history through hierarchical storage, temporal updates, structured retrieval, or latent memory~\citep{packer2023memgpt,zhong2024memorybank,sarthi2024raptor,gutierrez2024hipporag,xu2025amem,zhang2026memgen}. EgoMem and VoiceMem connect persistent memory to streaming multimodal interaction~\citep{yao2025egomem,xie2026voicemem}. These methods improve retention and retrieval but process session-level evidence after observation; they do not jointly control event admission, causal supersession, or exclusion of non-addressed and incomplete speech during concurrent interaction.

\textbf{Full-Duplex and Audio Benchmarks.}
Full-Duplex-Bench evaluates pauses, backchannels, turn transitions, and interruptions; its overlap-focused extension adds speech to others and background speech~\citep{lin2025fullduplexbench,lin2026fullduplexbenchv15}. HumDial-FDBench broadens human-recorded coverage, while adaptive benchmarks test correction handling, entity tracking, tool use, and interruption-conditioned state updates~\citep{wang2026humdial,lin2026fullduplexbenchv2,lin2026fullduplexbenchv3,modi2026echochain}. General audio benchmarks assess speech, scene, and paralinguistic understanding~\citep{wang2025audiobench}; Audio MultiChallenge also evaluates multi-turn inference memory, instruction retention, coherence, and spoken self-correction~\citep{gosai2026audiomultichallenge}. These protocols broaden audio-interaction evaluation but do not expose an event-level persistent-state ledger under adaptive full-duplex execution.


\begin{table}[t!]
\centering
\begin{threeparttable}
\caption{\textsf{DuplexLedger} versus representative full-duplex and conversational-memory benchmarks.}
\label{tab:fd_dataset_comparison}

\scriptsize
\setlength{\tabcolsep}{3.5pt}
\renewcommand{\arraystretch}{1.10}

\begin{tabular*}{\textwidth}{@{\extracolsep{\fill}}lrrcccccc@{}}
\toprule
& & & \multicolumn{2}{c}{\textbf{Duplex}} & \multicolumn{4}{c}{\textbf{Memory}} \\
\cmidrule(lr){4-5}\cmidrule(lr){6-9}
\textbf{Dataset} & \textbf{Scale} & \textbf{Hrs} &
\shortstack{\textbf{Adaptive}\\\textbf{interaction}} &
\shortstack{\textbf{FD}\\\textbf{scenarios}} &
\shortstack{\textbf{State / Memory}\\\textbf{eval.}} &
\shortstack{\textbf{Update}\\\textbf{QA}} &
\shortstack{\textbf{Distractor}\\\textbf{reject}} &
\shortstack{\textbf{Event}\\\textbf{state}} \\
\midrule
FDB v1/v1.5       & 727/499  & NR       & \xmark & 4/4 & \xmark & \xmark & \xmark & \xmark \\
HumDial-FDBench   & 15{,}698 & $>100$   & \xmark & 9   & \xmark & \xmark & \xmark & \xmark \\
MultiTalkBench    & 104      & 56.5     & \xmark & N/A & \cmark & \xmark & \xmark & \xmark \\
FDB v2            & 200      & N/A      & \cmark & 4   & \cmark & \xmark & \xmark & \xmark \\
EchoChain         & 200      & NR       & \cmark & 1   & \cmark & \xmark & \xmark & \xmark \\
\midrule
LoCoMo            & 10       & N/A      & \xmark & N/A & \cmark & \xmark & \xmark & \xmark \\
LongMemEval       & 500      & N/A      & \xmark & N/A & \cmark & \cmark & \xmark & \xmark \\
\midrule
\rowcolor{OursBlue}
\textbf{\textsf{DuplexLedger} (ours)} & \textbf{1{,}094} & \textbf{$\boldsymbol{\geq 35.3}$}
& \cmark & \textbf{4} & \cmark & \cmark & \cmark & \cmark \\
\bottomrule
\end{tabular*}

\begin{tablenotes}[flushleft]
\scriptsize
\item[] Scale denotes stimuli for FDB v1/v1.5, instances for HumDial-FDBench,
samples for MultiTalkBench, tasks for FDB v2, interrupted conversations for
EchoChain, conversations for LoCoMo, questions for LongMemEval, and sessions
for \textsf{DuplexLedger}. \textit{Update QA} denotes explicit evaluation of factual
revision or knowledge updating; temporal reasoning alone is not counted.
NR: aggregate audio duration is not reported; N/A: no fixed audio corpus or
no comparable full-duplex scenario taxonomy.
\end{tablenotes}
\end{threeparttable}
\end{table}

\section{DuplexLedger: Benchmark Design}
\label{sec:duplexledger}

Full-duplex memory evaluation requires determining event content and whether it should update persistent state. \textsf{DuplexLedger} combines adaptive interaction challenges with event-level memory supervision, enabling joint evaluation of interaction behavior and memory fidelity.

\begin{figure}[t!]
    \centering
    \includegraphics[width=\linewidth]{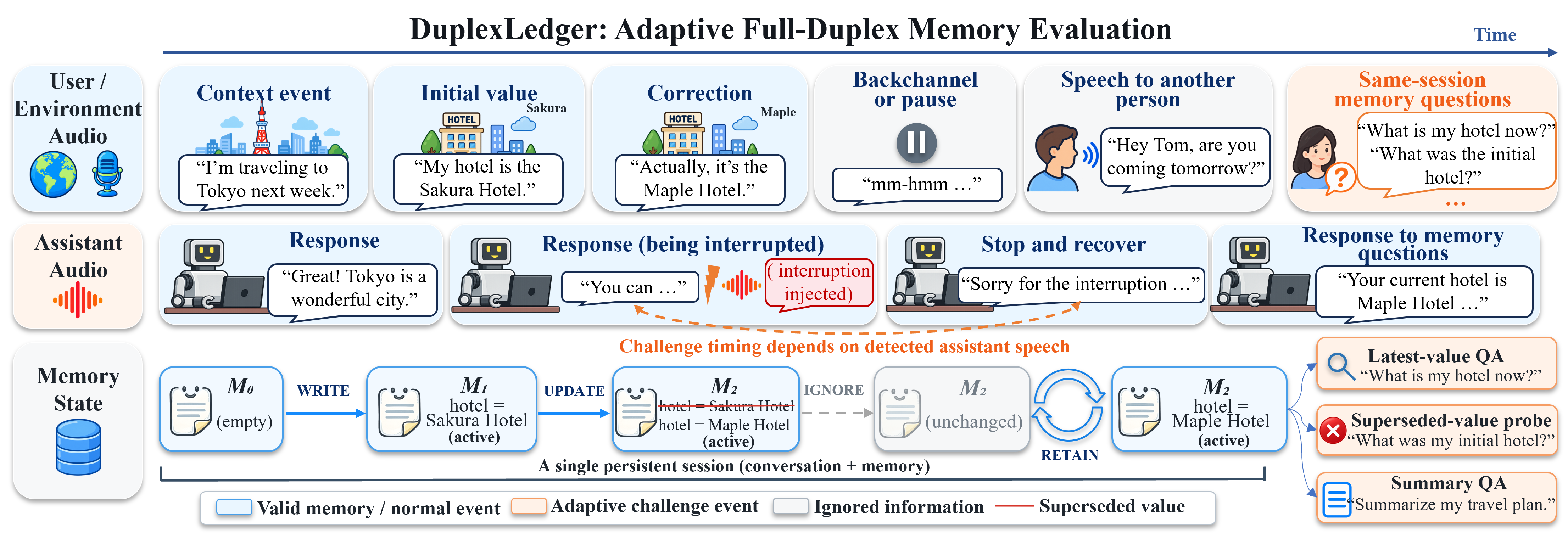}
    \caption{Overview of \textsf{DuplexLedger}. A persistent session aligns adaptively triggered full-duplex challenges, event-level memory transitions, and spoken memory questions.}
    \label{fig:duplexledger_overview}
\end{figure}

\subsection{Task Formulation}
\label{sec:duplexledger_formulation}

A session is an ordered event stream $\mathcal{E}=(e_1,\ldots,e_T)$, where $e_t=(x_t,s_t,a_t,\tau_t)$ specifies speech content, speaker, addressee, and timing. The memory state $M_t$ contains items $(k,v,\sigma)$ with entity key $k$, value $v$, and status $\sigma\in\{\textit{active},\textit{superseded}\}$. Because one event may affect multiple entities, transitions are defined at the item level:
\begin{equation}
    M_t=\mathcal{T}\!\left(M_{t-1},\,\{(k_i,v_i,o_i)\}_{i=1}^{n_t}\right),
    \qquad
    o_i\in\{\textsc{Write},\textsc{Update},\textsc{Ignore},\textsc{Retain}\}.
    \label{eq:duplexledger_transition}
\end{equation}
\textsc{Write} introduces a value, \textsc{Update} supersedes an active value, \textsc{Ignore} rejects inadmissible evidence, and \textsc{Retain} preserves the current state. The evaluated model receives only streaming audio and spoken questions; ledger operations and states are reserved for supervision and scoring. The task therefore evaluates selective state construction rather than unrestricted recall.

\subsection{Adaptive Full-Duplex Sessions}
\label{sec:duplexledger_sessions}

Each persistent session contains contextual events, two to five challenges, and six evaluation targets, without state resets. Fixed event content follows a model-dependent schedule: interruptions and backchannels require VAD-detected decoded assistant speech, with unmet triggers remaining unrealized. Pause and speech-to-others events use predefined observation windows.

The four scenarios require distinct behavior. An interruption should stop the current response and admit the revised intent; a backchannel should neither terminate the response nor enter memory. A within-turn pause should not trigger premature generation or commitment, whereas speech addressed to another person should produce neither an assistant response nor a memory update.

\subsection{Event-Level Supervision and Evidence-Grounded QA}
\label{sec:duplexledger_supervision}

Each ledger entry records the event content and roles, target operation, memory-state transition, and supersession links. Each question is aligned with supporting memory identifiers, forbidden superseded or distractor identifiers, and a reference answer. Two latest-value questions per session test whether revisions replace earlier values; the remaining targets cover factual recall, distractor rejection, final-state summarization, and scenario behavior.

Combined with system traces, this alignment supports diagnosis of admission, supersession, exclusion, retrieval, and generation failures. Table~\ref{tab:duplexledger_statistics} summarizes dataset composition and supervision.

\begin{table}[t]
\centering
\caption{Composition and supervision statistics of \textsf{DuplexLedger}.}
\label{tab:duplexledger_statistics}
\scriptsize
\renewcommand{\arraystretch}{1.10}

\begin{minipage}[t]{0.535\linewidth}
\vspace{0pt}
\centering
\textbf{(a) Session composition}\par\vspace{0.7mm}
\setlength{\tabcolsep}{1.8pt}
\begin{tabular*}{\linewidth}{@{\extracolsep{\fill}}lrrrrrr@{}}
\toprule
\textbf{Scenario} & \textbf{ZH} & \textbf{EN} & \textbf{Sess.} & \textbf{Events} & \textbf{Chal.} & \textbf{QA} \\
\midrule
Pause handling    & 149 & 127 & 276 & 1,932 &   552 & 1,656 \\
Speech to others  & 150 & 126 & 276 & 2,208 & 1,104 & 1,656 \\
User backchannel  & 146 & 137 & 283 & 2,545 & 1,130 & 1,698 \\
User interruption & 132 & 127 & 259 & 1,554 &   518 & 1,554 \\
\midrule
\rowcolor{LedgerBlue}
\textbf{Total} & \textbf{577} & \textbf{517} & \textbf{1,094} & \textbf{8,239} & \textbf{3,304} & \textbf{6,564} \\
\bottomrule
\end{tabular*}
\end{minipage}\hfill
\begin{minipage}[t]{0.445\linewidth}
\vspace{0pt}
\centering
\textbf{(b) Ledger and evaluation supervision}\par\vspace{0.7mm}
\setlength{\tabcolsep}{1.5pt}
\begin{tabular*}{\linewidth}{@{\extracolsep{\fill}}lrr@{\hspace{5pt}}lrr@{}}
\toprule
\textbf{Operation} & \textbf{Count} & \textbf{\%} & \textbf{Target} & \textbf{Count} & \textbf{/sess.} \\
\midrule
\textsc{Write}  & 3,982 & 27.1 & Latest value       & 2,188 & 2 \\
\textsc{Update} & 1,212 &  8.3 & Factual recall     & 1,094 & 1 \\
\textsc{Retain} & 1,941 & 13.2 & Distractor reject. & 1,094 & 1 \\
\textsc{Ignore} & 7,546 & 51.4 & Final summary      & 1,094 & 1 \\
\midrule
\cellcolor{LedgerBlue}\textbf{Total}
& \cellcolor{LedgerBlue}\textbf{14,681}
& \cellcolor{LedgerBlue}\textbf{100}
& Behavior check & 1,094 & 1 \\
\bottomrule
\end{tabular*}
\end{minipage}

\vspace{1mm}
\parbox{\linewidth}{\raggedright\scriptsize
\textit{Notes.} Events are executable user or bystander inputs; ledger counts additionally include 6,442 aligned assistant-shadow entries. Fixed-input and event-source audio total 35.33 and 24.22\,h, respectively. Release dev/test splits contain 237/857 sessions; method train/validation/calibration/held-out splits contain 660/175/40/219 sessions.}
\end{table}

\subsection{Construction and Evaluation Protocol}
\label{sec:duplexledger_protocol}

\textsf{DuplexLedger} comprises controlled bilingual sessions generated specifically for this benchmark. We preserve source waveforms, derive event playback windows from timestamps and forced-alignment boundaries, and compile immutable manifests specifying event order, response requirements, challenge conditions, and same-session questions. Gold operations, memory states, and reference answers are retained for supervision, scoring, and audit, but withheld from the evaluated speech model and its answer generator. Appendix~\ref{app:duplexledger_construction} details construction, quality controls, and the data card.Adaptive evaluation distinguishes planned from realized challenges. Interruption and backchannel events are triggered by decoded assistant speech, whereas pause and speech-to-others conditions use scheduled observation windows. For session $s$, the realized challenge rate (RCR) and its macro-average are defined as
\begin{equation}
\mathrm{RCR}_s =
\frac{N_{\mathrm{realized}}^{(s)}}
     {N_{\mathrm{planned}}^{(s)}},
\qquad
\overline{\mathrm{RCR}} =
\frac{1}{|\mathcal{S}|}
\sum_{s \in \mathcal{S}} \mathrm{RCR}_s,
\label{eq:rcr}
\end{equation}
where $\mathcal{S}$ contains sessions with at least one planned challenge.

Other session-level ratios are also macro-averaged unless stated otherwise. Conditional interaction metrics use eligible realized or observable events and are reported alongside RCR and validity indicators. All memory answers are scored by the same deterministic bilingual lexical scorer. Memory questions retain fixed question denominators, so silence and unanswered cases receive no credit. Latest-value answers must assert the active value without asserting a superseded alternative, and silence does not count as distractor rejection. Exact scheduling rules and metric denominators are specified in Appendices~\ref{app:execution_protocol} and~\ref{app:metrics}; the released artifact structure is documented in Appendix~\ref{app:artifacts}.

\providecolor{FindingBlue}{HTML}{1F4E79}

\providecommand{\findingstatement}[2]{%
  \par\noindent
  \textcolor{FindingBlue}{\textbf{\textit{[Takeaway #1]}}}%
  \hspace{0.35em}\textbf{\textit{#2}}\hspace{0.35em}%
}

\section{Performance of Current Full-Duplex Models}
\label{sec:baseline_performance}

\subsection{Experimental Setup and Reporting Protocol}
\label{sec:baseline_setup}

We evaluate five full-duplex systems using identical source audio, event order, questions, and scheduler logic. Since model architectures and response tendencies differ, cross-model results are descriptive. Memory questions retain fixed denominators, while conditional interaction metrics follow their eligible-event denominators. RCR is therefore treated as challenge exposure rather than a performance score. The execution and scoring protocols are detailed in Appendices~\ref{app:execution_protocol} and~\ref{app:metrics}; model-specific inference settings and complete baseline results appear in Appendices~\ref{app:baseline_details} and~\ref{app:complete_results}.

\subsection{Benchmark Results}
\label{sec:benchmark_results}

\findingstatement{1}{No model jointly dominates full-duplex interaction and memory fidelity.}
Table~\ref{tab:baseline_results} separates the two capabilities. \texttt{Lychee-FD} leads in conditional full-duplex performance and latest-value accuracy. \texttt{MiniCPM}-o~4.5 achieves the lowest early-onset and pre-end overlap rates, together with the strongest distractor rejection and final-state coverage. \texttt{Moshi} pairs the lowest stale rate with only 54.31\% response coverage. \texttt{PersonaPlex} realizes the most challenges but obtains the lowest cFD. These contrasting profiles preclude a meaningful scalar ranking.

\begin{table}[t]
\centering
\begin{threeparttable}
\caption{Full-duplex interaction and memory fidelity on \textsf{DuplexLedger}.}
\label{tab:baseline_results}
\scriptsize
\setlength{\tabcolsep}{2.5pt}
\renewcommand{\arraystretch}{1.10}

\begin{tabularx}{\textwidth}{@{}l*{7}{>{\centering\arraybackslash}X}@{}}
\toprule
\rowcolor{FDHeader}
\multicolumn{8}{c}{\textbf{(a) Full-duplex interaction}} \\
\midrule
\textbf{Model} &
\textbf{RCR}$^{\dagger}$ &
\textbf{cFD}$\uparrow$ &
\textbf{Early}$\downarrow$ &
\textbf{Pre-end}$\downarrow$ &
\textbf{Win-Intr.}$\downarrow$ &
\textbf{Stop@0.8}$\uparrow$ &
\textbf{Stop (s)}$\downarrow$ \\
\midrule
\texttt{MiniCPM}-o~4.5
& 73.51 & \underline{40.23} & \textbf{18.70} & \textbf{44.38}
& 59.47 & 55.69 & 1.032 \\
\rowcolor{FDStripe}
\texttt{Freeze-Omni}
& 86.17 & 28.40 & 49.76 & 91.82
& 73.64 & 45.75 & 1.239 \\
\texttt{Moshi}
& 44.63 & 37.36 & 66.59 & 72.25
& \textbf{45.11} & \underline{60.79} & \underline{0.777} \\
\rowcolor{FDStripe}
\texttt{Lychee-FD}
& 75.64 & \textbf{47.32} & \underline{24.68} & \underline{57.14}
& \underline{53.89} & \textbf{90.12} & \textbf{0.312} \\
\texttt{PersonaPlex}
& 93.54 & 22.74 & 46.64 & 72.73
& 90.62 & 16.40 & 1.689 \\
\bottomrule
\end{tabularx}

\vspace{1mm}

\begin{tabularx}{\textwidth}{@{}l*{6}{>{\centering\arraybackslash}X}@{}}
\toprule
\rowcolor{FDHeader}
\multicolumn{7}{c}{\textbf{(b) Memory fidelity}} \\
\midrule
\textbf{Model} &
\textbf{Resp.}$\uparrow$ &
\textbf{Latest}$\uparrow$ &
\textbf{Stale}$\downarrow$ &
\textbf{D-Adopt}$\downarrow$ &
\textbf{D-Rej.}$\uparrow$ &
\textbf{Final}$\uparrow$ \\
\midrule
\texttt{MiniCPM}-o~4.5
& \underline{98.90} & 4.30 & \underline{0.59}
& \textbf{14.17} & \textbf{85.19} & \textbf{2.01} \\
\rowcolor{FDStripe}
\texttt{Freeze-Omni}
& \textbf{99.94} & \underline{11.61} & 3.38
& 65.90 & 34.10 & 0.00 \\
\texttt{Moshi}
& 54.31 & 2.19 & \textbf{0.32}
& \underline{44.15} & 8.32 & 0.04 \\
\rowcolor{FDStripe}
\texttt{Lychee-FD}
& 92.41 & \textbf{25.50} & 1.65
& 48.17 & \underline{51.55} & \underline{1.06} \\
\texttt{PersonaPlex}
& 96.40 & 5.85 & 2.29
& 79.71 & 17.46 & 0.24 \\
\bottomrule
\end{tabularx}

\begin{tablenotes}[flushleft]
\scriptsize\raggedright
\item[] \textbf{Bold} and \underline{underlined}: best and second best. Values are percentages except Stop (s). $\dagger$ RCR is diagnostic; cFD is conditional. Win-Intr.: windowed speech intrusion; D-Adopt/D-Rej.: distractor adoption/rejection; Final: final-state coverage.
\end{tablenotes}
\end{threeparttable}
\end{table}

\subsection{Full-Duplex Coordination under Adaptive Challenges}
\label{sec:fd_coordination}

\findingstatement{2}{Challenge exposure, boundary control, and interruption recovery are distinct capabilities.}
\texttt{Lychee-FD} combines substantial exposure with the highest cFD, 90.12\% stopping success within 800\,ms, and the shortest stopping latency. Its remaining pre-end overlap and window intrusion nevertheless show that fast interruption recovery does not ensure correct turn boundaries. \texttt{MiniCPM}-o~4.5 instead minimizes early onset and pre-end overlap, while \texttt{PersonaPlex} pairs the highest RCR with the lowest cFD, greatest window intrusion, and slowest stopping. RCR must therefore be interpreted as exposure rather than interaction quality.

\subsection{Transactional Memory Failures and Interaction--Memory Misalignment}
\label{sec:transactional_failures}

\findingstatement{3}{Local value revision does not yield a coherent persistent state.}
\texttt{Lychee-FD} reaches 25.50\% latest-value accuracy but covers only 1.06\% of final-state facts and adopts 48.17\% of distractors. \texttt{MiniCPM}-o~4.5 is more selective but correctly answers only 4.30\% of update questions. \texttt{Freeze-Omni} responds almost universally but retains high stale-value and distractor errors. \texttt{Moshi} similarly combines a low stale rate with limited response coverage and latest-value accuracy.

Even the best final-state coverage is only 2.01\%. Current systems can coordinate speech, recover isolated revisions, or reject irrelevant content, yet none jointly supports event admission, supersession, exclusion, and state consolidation. Maintaining an evolving state during full-duplex interaction is therefore the central limitation motivating \textsf{ACT-Mem} in Section~\ref{sec:actmem}.

\section{ACT-Mem: Transactional Memory for Full-Duplex Agents}
\label{sec:actmem}

\subsection{Design Motivation and Overview}

\begin{figure}[t]
    \centering
    \includegraphics[width=\linewidth]{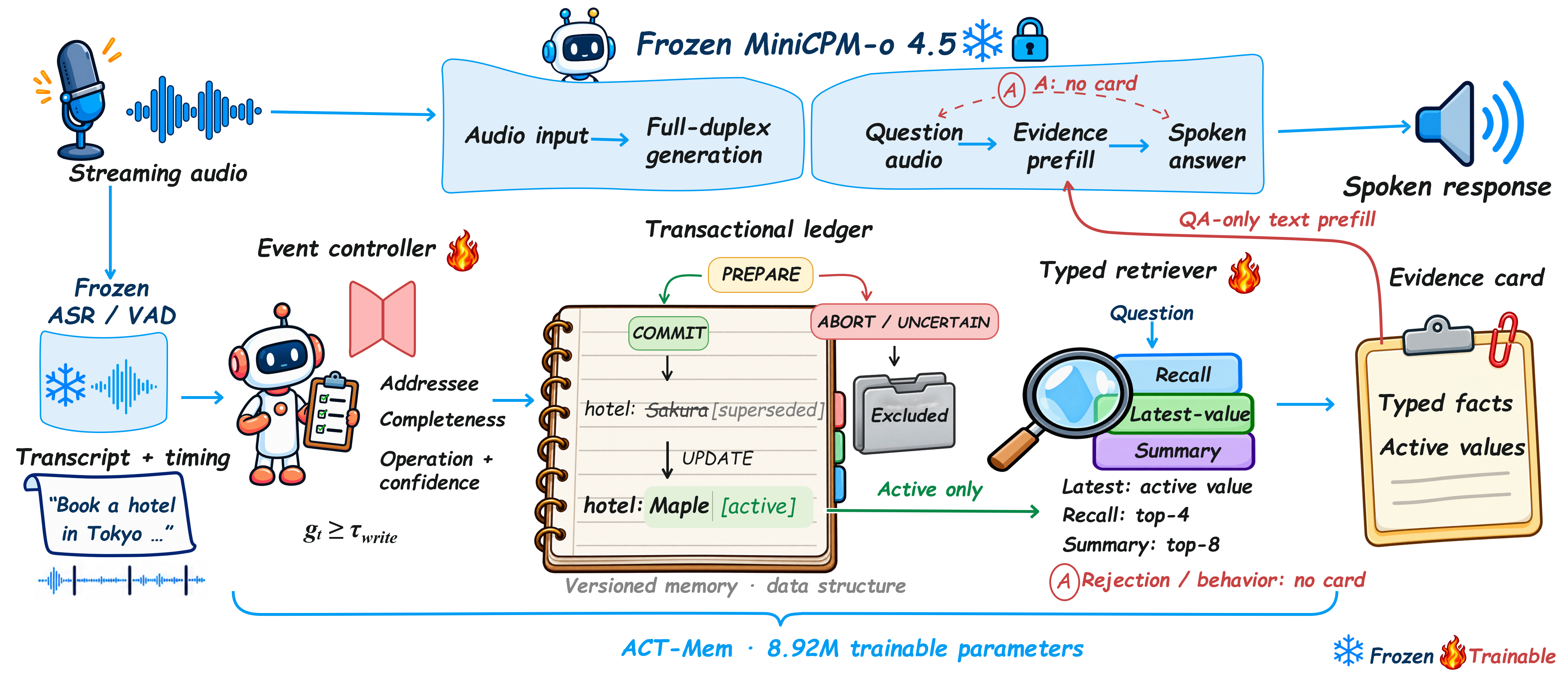}
    \caption{Overview of the \textsf{ACT-Mem} framework for transactional memory in full-duplex agents.}
    \label{fig:actmem_architecture}
\end{figure}

The failures in Section~\ref{sec:transactional_failures} motivate explicit control over memory admission and revision. We introduce \textsf{ACT-Mem} (\textbf{A}uditable and \textbf{C}ausal \textbf{T}ransactional Memory), an external module coupled with a frozen \texttt{MiniCPM}-o~4.5 backbone~\citep{minicpmo45}. As shown in Figure~\ref{fig:actmem_architecture}, streaming audio enters both the frozen full-duplex backbone and a frozen ASR/VAD module. The latter supplies transcripts and timing to a trainable event controller, which manages a transactional ledger implemented as versioned memory. A typed retriever constructs evidence cards for memory questions. Only these cards enter the backbone through QA-only text prefill; all speech-generation components remain frozen. Architectural and runtime details are provided in Appendix~\ref{app:actmem_implementation}.

\subsection{Event Admission and Transaction Lifecycle}

\methodinnovation{Admission}{Persistence requires explicit evidence admission.}
A deterministic parser extracts candidate entity--value slots from the event transcript, while the learned controller predicts event-level addressee, completeness, operation, and confidence for admission. Each proposal enters \textsc{Prepare} with score
\begin{equation}
    g_t =
    p_t^{\mathrm{addr}}p_t^{\mathrm{complete}}
    \bigl[p_t(\textsc{Write})+p_t(\textsc{Update})\bigr]c_t,
    \label{eq:actmem_admission}
\end{equation}
where $p_t^{\mathrm{addr}}$ and $p_t^{\mathrm{complete}}$ estimate assistant-directedness and semantic completeness, and $c_t\in[0,1]$ is the controller confidence.

A proposal enters \textsc{Commit} when $g_t\geq\tau_{\mathrm{write}}$. Non-addressed or incomplete proposals enter \textsc{Abort}; ambiguous or low-confidence proposals remain \textsc{Uncertain}. Neither is retrievable. Pause and silence create no persistent transaction. \textsc{Write} creates an active fact, \textsc{Update} replaces an active value, \textsc{Ignore} rejects the proposal, and \textsc{Retain} preserves the ledger. Committed entries retain their source event, predicted roles, order, completeness, and confidence for audit.

\subsection{Causal Supersession and Versioned Memory}

\methodinnovation{Supersession}{Recency alone cannot overwrite an active value.}
The transactional ledger extends the state in Section~\ref{sec:duplexledger_formulation} with provenance and version links. A candidate $m_j$ may supersede an active item $m_i$ only when
\begin{equation}
    k_j=k_i,\qquad
    t_j>t_i,\qquad
    \widehat{o}_j=\textsc{Update},\qquad
    g_j\geq\tau_{\mathrm{write}},
    \label{eq:actmem_supersession}
\end{equation}
where $k_i$ is the entity key and $t_i$ denotes source-event order. A later same-key mention is therefore insufficient without an admissible update.

A successful transaction marks $m_i$ as \textit{superseded}, activates $m_j$, and records their correction link. Both versions remain auditable, but only active values are exposed to the typed retriever. Superseded, uncertain, aborted, and ignored content cannot enter an evidence card.

\subsection{Typed Retrieval and Memory Conditioning}

\methodinnovation{Typed retrieval}{Question type controls evidence scope and card generation.}
The typed retriever maps each question to one of four behaviors. A \emph{latest-value} question selects the target entity's active value, a \emph{recall} question retrieves up to four active facts, and a \emph{summary} question aggregates up to eight facts across types and entities. Rejection and behavior questions produce no card, preventing inadmissible content from being disclosed to the backbone.

For retrieval-enabled questions, ranking combines dense similarity~\citep{karpukhin2020dpr}, lexical overlap, recency, controller confidence, and stale/distractor penalties. The resulting card contains only typed facts and active values; insufficient evidence yields no card.

\textsf{ACT-Mem} conditions generation without modifying the decoder~\citep{lewis2020rag}. At a stable-silence QA boundary, the backbone processes the question audio, receives one evidence card through QA-only text prefill, and generates a spoken answer. Context events and full-duplex challenges continue through the native audio interface without memory prefill.

\subsection{Training, Calibration, and Model Complexity}

Training uses frozen-ASR transcripts from the session-disjoint training split. The event controller learns operation, addressee, completeness, and confidence prediction. Slot extraction and entity-key normalization use non-trainable rules, and the ledger applies version transitions. Its objective is
\begin{equation}
\begin{aligned}
\mathcal{L}_{\mathrm{ctrl}}
&=\mathcal{L}_{\mathrm{op}}^{\mathrm{wCE}}
+0.3\mathcal{L}_{\mathrm{addr}}^{\mathrm{CE}}
+0.3\mathcal{L}_{\mathrm{complete}}^{\mathrm{CE}}
+0.2\mathcal{L}_{\mathrm{conf}}^{\mathrm{BCE}}, \\
\mathcal{L}_{\mathrm{retr}}
&=\mathcal{L}_{\mathrm{BCE}}(\mathbf{s},\mathbf{y})
+\frac{0.5}{|\mathcal{P}||\mathcal{N}|}
\sum_{p\in\mathcal{P}}\sum_{n\in\mathcal{N}}
\left[0.25-s_p+s_n\right]_{+}.
\end{aligned}
\label{eq:actmem_training}
\end{equation}
Weighted operation cross-entropy uses class weights $(1.0,1.5,1.4,0.8)$ for \textsc{Write}, \textsc{Update}, \textsc{Ignore}, and \textsc{Retain}. $\mathbf{s}$ contains retrieval scores, $\mathbf{y}$ gives binary relevance labels, and $\mathcal{P}$ and $\mathcal{N}$ contain positive and forbidden, hard-negative, or distractor items in the QA context. Joint training updates the controller and retriever; adaptation freezes the controller and updates only the retriever.

We use the session-disjoint splits in Section~\ref{sec:duplexledger_protocol}. Gold states and evidence labels provide supervision only; future events never enter the current prefix. Calibration on 40 sessions gives $\tau_{\mathrm{write}}=0.35$ without gradient updates. \textsf{ACT-Mem} adds 8.92M trainable parameters: 4.59M in the event controller and 4.33M in the typed retriever. The backbone and ASR/VAD module remain frozen. Training uses 8 NVIDIA A100 GPUs; optimization and augmentation details are provided in Appendix~\ref{app:training_details}.


\providecolor{ACTHeader}{HTML}{E8E7EE}
\providecolor{ACTStripe}{HTML}{F5F5F5}
\providecolor{ACTBackbone}{HTML}{F0F0F0}
\providecolor{ACTBlue}{HTML}{E6F0FB}
\providecolor{ACTImproveRed}{HTML}{D65245}
\providecolor{ACTDeclineCyan}{HTML}{1696A7}
\providecolor{ACTNeutral}{HTML}{666666}
\providecommand{\finding}[1]{\noindent\textbf{\textit{#1}}}

\providecommand{\actimp}[1]{\textcolor{ACTImproveRed}{\tiny #1}}
\providecommand{\actreg}[1]{\textcolor{ACTDeclineCyan}{\tiny #1}}
\providecommand{\actneutral}[1]{\textcolor{ACTNeutral}{\tiny #1}}

\section{Evaluation of ACT-Mem}
\label{sec:actmem_evaluation}

We evaluate \textsf{ACT-Mem} on the same 219 held-out sessions used for the baseline study, containing 1,314 questions/checks and 438 latest-value questions. The controlled comparison retains the frozen \texttt{MiniCPM}-o~4.5 backbone; no held-out session is used for training, checkpoint selection, or calibration. Evaluation follows Sections~\ref{sec:duplexledger_protocol} and~\ref{sec:baseline_setup}. Relative changes use unrounded scores, and pointwise 95\% confidence intervals for latest-value accuracy and final-state coverage are obtained from 20,000 paired bootstrap resamples of sessions. Further details appear in Appendices~\ref{app:metrics} and~\ref{app:statistics}.

\subsection{Main Results}

\begin{figure}[t!]
    \centering
    \includegraphics[width=\linewidth]{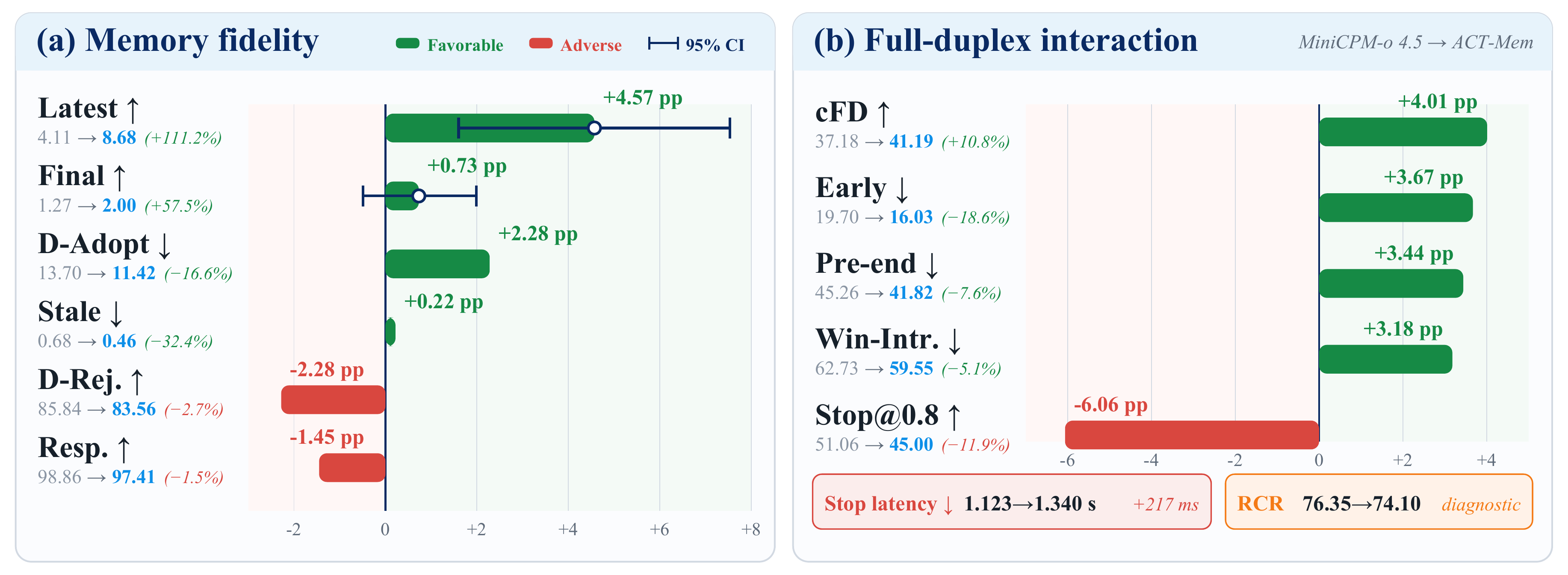}
    \vspace{-20pt}
    \caption{\textsf{ACT-Mem} versus its frozen \texttt{MiniCPM}-o~4.5 backbone on 219 held-out sessions; error bars show 95\% paired-bootstrap CIs for Latest and Final.}
    \vspace{-5pt}
    \label{fig:actmem_main_results}
\end{figure}

Figure~\ref{fig:actmem_main_results} compares the complete \textsf{ACT-Mem} configuration with its frozen \texttt{MiniCPM}-o~4.5 backbone.

\findingstatement{1}{\textsf{ACT-Mem} substantially improves latest-value recovery, while complete final-state reconstruction remains difficult.}
Latest-value accuracy increases from 4.11\% to 8.68\% 
($+111.2\%$; $+4.57$ percentage points, 95\% CI $[1.60, 7.53]$), 
whereas final-state coverage rises from 1.27\% to 2.00\% 
($+57.5\%$; $+0.73$ points, 95\% CI $[-0.49, 1.99]$). 
The latest-value interval excludes zero, providing evidence that \textsf{ACT-Mem} more reliably recovers active committed values after revision. Final-state coverage, however, remains low and its interval crosses zero. The controlled evidence therefore supports improved local state revision, but not yet complete reconstruction of the session-wide memory state.

\findingstatement{2}{The memory gains are concentrated in selective commitment rather than uniformly improving every outcome.}
Stale assertions and distractor adoption decrease by 0.22 and 2.28 percentage points, respectively, showing that the latest-value gain is accompanied by fewer conflicting and incorrectly admitted facts. Distractor rejection and response rate decline by 2.28 and 1.45 points, although the latter remains high at 97.41\%. Because distractor adoption and rejection are separately credited under fixed denominators, they are not complementary; their joint decline means that \textsf{ACT-Mem} produces fewer explicit adoptions but also fewer credited rejections. The principal benefit is therefore more selective state construction, while robust distractor rejection remains improvable.

\subsection{Effects on Full-Duplex Interaction}
\findingstatement{3}{\textsf{ACT-Mem} improves most interaction-boundary metrics but slows interruption stopping.}
cFD increases by 10.8\%, while early response, pre-end speech, and windowed intrusion decrease by 18.6\%, 7.6\%, and 5.1\%, respectively. Because cFD is conditioned on realized challenges and RCR decreases from 76.35\% to 74.10\%, this improvement should be interpreted as better behavior on exposed challenges rather than uniformly greater challenge coverage. The main trade-off is interruption stopping: Stop@0.8 falls from 51.06\% to 45.00\%, and median stopping latency increases from 1.123 to 1.340 seconds. Thus, \textsf{ACT-Mem} improves observed onset and overlap control, while rapid interruption recovery remains its principal interaction weakness.

\vspace{-5pt}
\subsection{Ablation Studies}
\label{sec:actmem_ablations}

We evaluate three inference-time variants with fixed memory-model weights. \emph{FIFO memory} replaces the versioned ledger with a 32-slot queue and removes supersession and latest-only selection. \emph{Without addressee gating} removes speaker/addressee rejection and the learned gate. \emph{Untyped retrieval} applies top-4 retrieval to every question type, including rejection questions.

\findingstatement{4}{The full configuration balances value recovery with selective memory use.}
\textsf{ACT-Mem} leads all three reported interaction metrics and three of four memory metrics. Compared with FIFO memory, its stale-assertion rate is 79.8\% lower; compared with removing addressee gating, distractor adoption is 34.2\% lower. Untyped retrieval achieves 9.59\% latest-value accuracy, but stale assertions and distractor adoption reach 1.60\% and 19.63\%. \textsf{ACT-Mem} thus trades 0.91 percentage points of latest-value accuracy for fewer stale assertions and distractor adoptions. Interaction differences remain descriptive across runs: QA-only retrieval does not establish a causal effect on preceding interaction boundaries.

\begin{table}[t]
\centering
\begin{threeparttable}
\caption{Ablations on 219 sessions. Best and second-best scores are bold and underlined.}
\label{tab:actmem_ablations}
\scriptsize
\setlength{\tabcolsep}{3pt}
\renewcommand{\arraystretch}{1.08}

\begin{tabularx}{\textwidth}{@{}l*{7}{>{\centering\arraybackslash}X}@{}}
\toprule
\rowcolor{ACTHeader}
& \multicolumn{3}{c}{\textbf{Full-duplex interaction}}
& \multicolumn{4}{c}{\textbf{Memory fidelity}} \\
\cmidrule(lr){2-4}\cmidrule(lr){5-8}
\rowcolor{ACTHeader}
\textbf{Variant} &
\textbf{cFD}$\uparrow$ &
\textbf{Early}$\downarrow$ &
\textbf{Pre-end}$\downarrow$ &
\textbf{Latest}$\uparrow$ &
\textbf{Stale}$\downarrow$ &
\textbf{D-Adopt}$\downarrow$ &
\textbf{Final}$\uparrow$ \\
\midrule

FIFO memory &
34.63 &
37.21 &
48.65 &
7.08 &
2.28 &
\underline{12.33} &
1.61 \\

\rowcolor{ACTStripe}
w/o addressee gating &
40.14 &
31.48 &
46.36 &
6.39 &
\underline{0.91} &
17.35 &
1.75 \\

Untyped retrieval &
\underline{40.33} &
\underline{29.98} &
\underline{44.61} &
\textbf{9.59} &
1.60 &
19.63 &
\underline{1.82} \\

\rowcolor{ACTBlue}
\textbf{\textsf{ACT-Mem}} &
\textbf{41.19} &
\textbf{16.03} &
\textbf{41.82} &
\underline{8.68} &
\textbf{0.46} &
\textbf{11.42} &
\textbf{2.00} \\

\bottomrule
\end{tabularx}

\begin{tablenotes}[flushleft]
\scriptsize\raggedright
\item[] All values are percentages. 
\end{tablenotes}
\end{threeparttable}
\end{table}

\subsection{Results by Language and Scenario}
\label{sec:actmem_group_analysis}
\findingstatement{5}{Improvements are bilingual but scenario-dependent.}
Latest-value accuracy increases from 5.83\% to 13.11\% on English sessions and from 2.59\% to 4.74\% on Chinese sessions. The largest absolute gains occur for backchannels (+10.53 points) and speech to others (+6.36), followed by pause handling (+1.82). Under user interruption, however, latest-value accuracy decreases from 3.85\% to 2.88\%; backchannel final-state coverage also falls from 2.34\% to 1.10\%. These descriptive breakdowns identify interruption-time updating as the principal remaining weakness.

\subsection{Case Study}
\label{sec:actmem_case_study}

\begin{figure}[t]
    \vspace{-6pt}
    \centering
    \includegraphics[width=\linewidth]{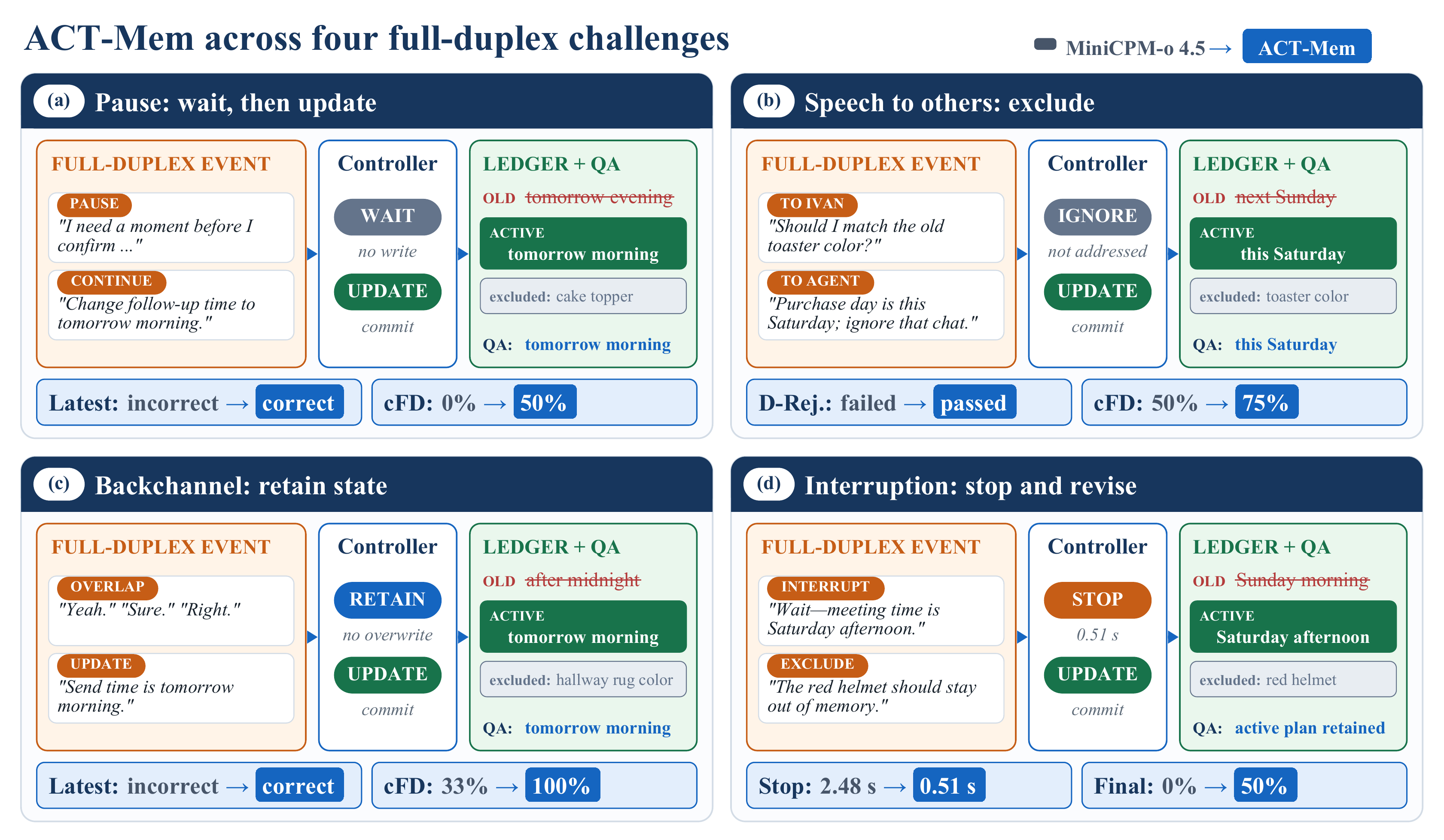}
    \vspace{-14pt}
    \caption{\textsf{ACT-Mem} cases across four full-duplex scenarios, showing controller decisions, memory transitions, and paired outcomes.}
    \vspace{-18pt}
    \label{fig:actmem_case_study}
\end{figure}

Figure~\ref{fig:actmem_case_study} connects interaction events with admission decisions, committed states, and QA outcomes. The controller governs admission; the ledger records supersession and exclusion; typed retrieval supplies active, question-relevant evidence. \textsf{DuplexLedger}'s reference transitions and \textsf{ACT-Mem}'s transaction traces distinguish correct operation predictions from successful slot resolution, supersession, or exclusion (Appendix~\ref{app:qualitative}, Table~\ref{tab:actmem_qualitative_trace}). These records support failure localization beyond answer scores by separating predicted operations, committed changes, and downstream QA outcomes.

\subsection{Admission-Threshold Sensitivity}
\label{sec:admission_threshold_sensitivity}

\providecolor{ACTHeader}{HTML}{E8E7EE}
\providecolor{ACTStripe}{HTML}{F5F5F5}

\setlength{\intextsep}{4pt}
\setlength{\columnsep}{10pt}

\begin{wraptable}{r}{0.60\textwidth}
\vspace{-4pt}
\centering
\small
\setlength{\tabcolsep}{3pt}
\renewcommand{\arraystretch}{1.05}
\setlength{\abovecaptionskip}{0pt}
\setlength{\belowcaptionskip}{0pt}

\caption{Threshold sensitivity on 40 calibration sessions.}
\label{tab:admission_threshold_sensitivity}

\vspace{2pt}

\begin{tabularx}{\linewidth}
{@{}l*{6}{>{\raggedleft\arraybackslash}X}@{}}
\toprule
\rowcolor{ACTHeader}
$\tau_{\mathrm{write}}$
& 0.35 & 0.45 & 0.50 & 0.55 & 0.65 & 0.75 \\
\midrule
Recall (\%)
& 55.26 & 55.26 & 54.74 & 53.68 & 50.53 & 48.95 \\
\rowcolor{ACTStripe}
F1 (\%)
& 71.19 & 71.19 & 70.75 & 69.86 & 67.13 & 65.72 \\
\bottomrule
\end{tabularx}

\vspace{-3pt}
\end{wraptable}

Table~\ref{tab:admission_threshold_sensitivity} summarizes admission sensitivity on 40 calibration sessions with fixed model weights. Event-level admission precision remains 100\% at all tested thresholds; 0.35 and 0.45 share the highest F1. Increasing the threshold from 0.35 to 0.75 reduces recall from 55.26\% to 48.95\% and F1 from 71.19\% to 65.72\%. These results indicate local stability across two adjacent settings and a recall cost from stricter admission, with conclusions limited to calibration-set admission behavior.
\par

\section{Conclusion}
\label{sec:conclusion}

We frame persistent memory during concurrent speech as selective state management. \textsf{DuplexLedger} aligns adaptive challenges, event-level operations, explicit reference states, and spoken QA across 1,094 bilingual sessions. This alignment supports joint evaluation of interaction and memory, while reference transitions and system traces help diagnose admission, supersession, exclusion, retrieval, and generation failures. Across five systems, no model dominates both interaction and memory fidelity. \textsf{ACT-Mem} combines an event controller, versioned ledger, and typed retriever with a frozen backbone. On 219 held-out sessions, latest-value accuracy rises from 4.11\% to 8.68\%, with fewer stale assertions and distractor adoptions. Several interaction-boundary metrics also improve, although interruption stopping becomes slower. These findings motivate explicit, auditable state management and further validation on natural conversations, additional languages, longer timescales, and other backbones.

\section*{Ethics Statement}
\textsf{DuplexLedger} contains AI-generated dialogues and synthetic speech rather than recordings of human participants or private conversations. Its controlled scenarios may nevertheless encode generation-model biases and should not be treated as representative of unrestricted real-world dialogue.

\section*{Reproducibility Statement}
Dataset construction, execution protocols, metric denominators, baseline configurations, \textsf{ACT-Mem} implementation and training, statistical procedures, complete baseline results, qualitative traces, and artifact organization are documented in Appendices~\ref{app:duplexledger_construction}--\ref{app:artifacts}.

\section*{AI Use Statement}
Generative models were used to construct synthetic benchmark dialogues as documented in Appendix~\ref{app:duplexledger_construction}. The authors specified the schemas, validation rules, evaluation protocol, and analyses, and remain responsible for the resulting content and claims.

\bibliography{iclr2027_conference}
\bibliographystyle{iclr2027_conference}

\appendix

\section{DuplexLedger Construction and Data Card}
\label{app:duplexledger_construction}

\subsection{Construction Principles}

\textsf{DuplexLedger} was constructed to evaluate whether a spoken dialogue agent can maintain a coherent memory state while simultaneously respecting full-duplex interaction boundaries. We designed each session around three coupled requirements: (i) the dialogue must contain facts that evolve through additions and corrections; (ii) some audible content must be inadmissible to persistent memory; and (iii) the interaction must contain a concurrency event whose correct handling depends on the model's runtime speaking state. These requirements led to four scenarios: user interruption, user backchannel, within-turn pause, and speech addressed to another person.

All dialogue content is synthetic. We used GPT-5.5 to generate the underlying Chinese and English task descriptions, conversational utterances, factual values, corrections, distractors, and reference answers under scenario-specific instructions. Generation was constrained by a structured schema specifying speaker roles, event order, memory dependencies, challenge types, and evaluation targets. The generated content was subsequently normalized and instantiated by deterministic construction programs. The language model was used only during dataset creation and is not part of the benchmark execution or scoring pipeline.

\subsection{Task and Dialogue Generation}

We constructed sessions from a broad collection of everyday task templates, including scheduling, planning, customer support, troubleshooting, study assistance, household coordination, travel, document preparation, creative work, and low-pressure personal organization. The Chinese and English construction pools contain 73 and 72 base task templates, respectively, covering 72 high-level domains in each language.

Each base task specifies two revisable attributes and several persistent facts. A typical task contains:

\begin{itemize}
    \item an initial value and a corrected value for a primary attribute;
    \item an initial value and a corrected value for a secondary attribute;
    \item a stable goal or constraint that should remain active;
    \item an additional task-specific detail;
    \item a named person and an associated responsibility;
    \item a plausible but irrelevant distractor;
    \item an expected consolidated final state.
\end{itemize}

To reduce exact repetition, each task was combined with a scenario-specific identifier, one of four discourse-style variants, and one of twelve instance-level modifiers. The modifiers vary properties such as budget conservatism, scheduling slack, communication requirements, fallback plans, execution complexity, and family or workplace coordination. Timing, challenge position, pause length, overlap duration, and short-feedback form were sampled using fixed construction seeds.

We also varied how the user concludes a session. A final turn may fully restate the task, omit the recap, provide only a partial recap, contain a self-correction, or explicitly request that the assistant not summarize. The intended proportions were 40\%, 25\%, 20\%, 10\%, and 5\%, respectively. The released corpus contains 434 complete recaps, 268 sessions without a complete final recap, 218 incomplete recaps, 118 recaps containing a correction, and 56 explicit no-summary requests. This variation prevents the final memory state from being recoverable solely by copying the last user turn.

\subsection{Scenario-Specific Interaction Design}

The same underlying memory requirements are combined with four distinct full-duplex phenomena.

\paragraph{User interruption.}
The user begins with a provisional task state and later interrupts an active assistant response to correct one or more values. The interruption should terminate the obsolete response trajectory, and the corrected values should supersede their earlier versions. A second interruption may clarify that a nearby detail is irrelevant and must not enter persistent memory.

\paragraph{User backchannel.}
Short acknowledgements such as ``yes,'' ``okay,'' or their Chinese equivalents are inserted while the assistant is speaking. These utterances provide listener feedback but introduce no new task fact. A correct system should continue its response without interpreting the backchannel as an interruption or memory write.

\paragraph{Pause handling.}
The user produces an incomplete utterance, pauses, and then resumes with a completion or correction. The pause is represented explicitly rather than as a completed turn. A correct system should wait for the continuation and must not commit a provisional value merely because a long silence occurred.

\paragraph{Speech addressed to another person.}
The session contains both user speech directed to a bystander and audible speech from that bystander. Such content is topically plausible but is not addressed to the assistant. It therefore serves as a controlled distractor: the model should neither respond to it nor incorporate it into the official task state. The user later returns explicitly to the assistant and continues the admissible task.

\subsection{Structured Event and Memory Annotation}

A session is represented as an ordered list of speech and silence events. Each event records its identifier, speaker, addressee, dialogue act, planned time interval, audio-rendering status, memory dependencies, and candidate memory writes. Events spoken by the primary user or bystander are audible; reference assistant events specify the intended conversational trajectory but are not rendered into the benchmark input.

Memory is annotated at the item level. The annotation vocabulary covers \texttt{goal}, \texttt{schedule}, \texttt{constraint}, \texttt{preference}, \texttt{episodic}, \texttt{named\_entity}, \texttt{task\_state}, and \texttt{correction}. Each item contains a stable memory identifier, semantic type, textual content, and source event. Corrections additionally identify the earlier item that they supersede. Distractors are annotated separately and are excluded from the oracle final state.

For evaluation, the event annotations are mapped to four memory operations: \textsc{Write}, \textsc{Update}, \textsc{Ignore}, and \textsc{Retain}. Ordinary admissible facts produce \textsc{Write}; explicit corrections and clarifications produce \textsc{Update}; backchannels, side conversations, and bystander speech produce \textsc{Ignore}; and pauses without confirmed content produce \textsc{Retain}. This mapping supplies the event-level supervision used to distinguish admission errors from supersession, retrieval, and answer-generation errors.

The construction procedure enforces two provenance constraints. First, every item in the oracle final state must be supported by an audible user event. Muted reference-assistant events cannot introduce a gold fact. Second, a superseding item must refer to an earlier item in the same session; future or cross-session references are disallowed.

\subsection{Memory Questions}

Every session contains six spoken evaluation targets:

\begin{enumerate}
    \item two latest-value questions, one for each corrected attribute;
    \item one factual-recall question about a stable requirement;
    \item one distractor-rejection question;
    \item one final-state summary question;
    \item one scenario-specific behavioral check.
\end{enumerate}

Each question is aligned with its expected answer, supporting memory identifiers or events, and, when applicable, forbidden obsolete or distractor items. The latest-value questions require the corrected value and prohibit credit when the obsolete value is asserted. The distractor question tests whether the system rejects inadmissible content rather than rewarding silence alone. The summary question requires coverage of multiple active facts and therefore evaluates state consolidation beyond isolated retrieval.

Questions are synthesized as audio and presented after all contextual and challenge events, without resetting the model state. The question audio, reference text, and gold evidence annotations are stored separately. Reference answers and memory annotations are used only for scoring and are not exposed to the evaluated system.

\subsection{Speech Synthesis and Temporal Alignment}

All speech is generated using text-to-speech synthesis; the dataset contains no recordings from human participants. Audible event segments are rendered at 24~kHz and assembled according to the structured event timeline. Scenario-specific speaker assignments distinguish the primary user from the bystander in multi-speaker sessions. The reference assistant remains muted, because the assistant audio during evaluation must be produced by the system under test.

We apply forced alignment to the synthesized speech to obtain word-level English boundaries and character-level Chinese boundaries. Punctuation is removed for alignment, while the original text is retained in the source annotation. The aligned event window is defined by the first and last aligned units with 120~ms of boundary padding. For each event, we retain the source sampling rate, number of samples, first and last aligned boundaries, leading and trailing silence, playback sample range, source duration, SHA-256 checksum, and quality-control flags.

The fixed mixed waveform is retained as a compatibility artifact, but it is not the authoritative input for adaptive evaluation. The final protocol instead replays aligned event segments according to a closed-loop stage manifest. This distinction is necessary because a full-duplex challenge should occur when the evaluated model is actually speaking, rather than at a timestamp derived from a prerecorded reference trajectory.

\subsection{Adaptive Session Materialization}

For each session, we compile a chronological execution manifest containing ordinary context playback, adaptive challenges, recovery periods, and same-session memory QA. Ordinary user events are streamed in real time. Before an interruption or backchannel, the scheduler waits for decoded-audio VAD to confirm that the evaluated model is speaking. It then injects the corresponding user event. If the required speaking state is not observed within the timeout, the challenge remains unrealized and the timeout is recorded.

Pause intervals are represented as explicit wait stages. Speech directed to another person is played through a dedicated side-talk stage and is marked as not addressed to the assistant. After an ordinary response-required event, the scheduler observes response onset and subsequent silence before advancing. The six audio questions are asked only after the interaction stages have finished.

The complete session uses one persistent model state:

\begin{itemize}
    \item the model is reset before a new session;
    \item it is not reset between context and challenges;
    \item it is not reset between challenges and memory QA;
    \item it is not reset between individual QA items.
\end{itemize}

The execution protocol uses 80-ms input chunks. Interruption and backchannel triggers require at least 500~ms and 800~ms of active assistant speech, respectively. Question input retains 400~ms of post-speech tail silence, and answer windows are determined dynamically. Runtime failures, trigger timeouts, and missing responses are recorded rather than silently discarded.

\subsection{Validation and Quality Control}

We apply validation at three levels.

\paragraph{Structural validation.}
Before synthesis, each session must contain all required fields, unique event and memory identifiers, chronologically valid timestamps, valid memory dependencies, and consistent overlap annotations. Memory reads may reference only previously introduced items. Supersession links must point to existing earlier items, and every final-state fact must trace to audible user speech.

\paragraph{Audio and alignment validation.}
The construction pipeline verifies the existence and duration of every waveform, preserves source-file checksums, and records missing alignments, leading silence, and trailing residuals. The released corpus contains no missing event-audio files. Of 7,687 speech events, 7,589 have forced-alignment boundaries. The remaining events are retained with explicit missing-alignment flags rather than removed. In addition, 3,007 aligned events contain more than 0.5~s of trailing residual audio; these cases remain visible through per-event quality-control metadata.

\paragraph{Execution validation.}
Each session must contain a nonempty ordered stage list, unique stage identifiers, relative and resolvable audio paths, an adaptive state-wait stage before every interruption or backchannel injection, and a final same-session QA stage. Question windows must satisfy
\[
0 \leq n_{\mathrm{start}}
< n_{\mathrm{speech\mbox{-}end}}
\leq n_{\mathrm{end}}.
\]

A full validation pass over all 1,094 sessions reports no hard structural errors.

\subsection{Dataset Composition}

Table~\ref{tab:duplexledger_data_card_appendix} summarizes the released corpus.

\begin{table}[t]
\centering
\caption{Composition of \textsf{DuplexLedger}. Fixed-input duration refers to the compatibility waveform; event-source duration is the total duration of the individual audible speech segments used by the adaptive protocol.}
\label{tab:duplexledger_data_card_appendix}
\scriptsize
\setlength{\tabcolsep}{4pt}
\renewcommand{\arraystretch}{1.10}

\begin{tabularx}{\textwidth}{@{}l>{\raggedleft\arraybackslash}X@{}}
\toprule
\textbf{Quantity} & \textbf{Value} \\
\midrule
Sessions & 1,094 \\
Chinese / English & 577 / 517 \\
Interruption / backchannel sessions & 259 / 283 \\
Pause / talking-to-other sessions & 276 / 276 \\
Development / test sessions & 237 / 857 \\
Memory questions & 6,564 \\
Latest-value questions & 2,188 \\
Full-duplex challenges & 3,304 \\
Audible speech and silence events & 8,239 \\
Speech-event waveforms & 7,687 \\
Fixed compatibility audio & 35.328 h \\
Event-source audio & 24.221 h \\
Session duration, mean / median & 116.25 s / 114.93 s \\
Session duration, minimum / maximum & 86.64 s / 176.70 s \\
MFA-aligned speech events & 7,589 \\
Missing event waveforms & 0 \\
\bottomrule
\end{tabularx}
\end{table}

The public release uses a fixed split of 237 development and 857 test sessions. The split assignment, source text, source waveforms, question order, and oracle annotations are held fixed in the adaptive release. The separate method experiments use session-disjoint training, validation, calibration, and held-out subsets; no persistent session is shared across those subsets.

\subsection{Intended Use and Access Restrictions}

The baseline setting of \textsf{DuplexLedger} evaluates spoken agents
through streamed audio events and spoken questions within a persistent
session. Source text, annotation timestamps, event roles, gold memory
operations, oracle states, evidence links, and reference answers are
evaluator-side resources rather than baseline-model inputs. For
augmented systems, access by auxiliary modules must be distinguished
from access by the speech backbone; ACT-Mem's observation and
conditioning interfaces are described in Appendix E.

Access to gold transcripts, annotated speaker/addressee labels,
question types, oracle states, or supporting-memory identifiers
must be declared for the complete evaluated system, including
auxiliary modules. Such access constitutes an annotation-assisted
or oracle setting, depending on the information supplied, and
must be distinguished from end-to-end audio-only evaluation.

\subsection{Limitations and Responsible Release}

\textsf{DuplexLedger} is fully synthetic and therefore does not reproduce the complete distribution of spontaneous human speech. In particular, TTS speech underrepresents natural disfluency, accent variation, emotional prosody, microphone conditions, environmental noise, and unplanned multi-party interaction. The sessions last approximately one to three minutes and do not establish performance on hour-scale or cross-day memory.

Although the task includes selective admission, correction, and distractor rejection, it does not currently contain an explicit memory-deletion operation. It should therefore not be interpreted as a complete benchmark for privacy-aware memory management. The absence of real-user recordings reduces direct privacy risk, but public release still requires documentation of the generation-model and speech-synthesis licenses, the exact generation prompts, construction seeds, source and derived checksums, and all validation reports.

\section{Adaptive Full-Duplex Execution Protocol}
\label{app:execution_protocol}

\subsection{Protocol Objective and Execution Unit}

The evaluation unit is a complete interaction session associated with one \textsf{DuplexLedger} sample. A session contains an annotated context, one or more full-duplex challenges, a recovery phase, and a set of memory questions. The audio content and event annotations are fixed before evaluation, whereas the wall-clock time at which a challenge is delivered is determined online from the decoded assistant waveform. Consequently, the protocol distinguishes between a \emph{planned} challenge, a \emph{trigger-satisfied} challenge, and a \emph{realized} challenge. A planned challenge is an event specified in the sample manifest. A trigger is satisfied only when the required assistant state is observed within the prescribed timeout. A challenge is realized only when the corresponding user audio is subsequently delivered to the model. This separation prevents a silent or non-responsive model from being credited with having successfully experienced an interruption that was never actually applied.

Each sample is evaluated in one persistent model session. The session is reset before the first context event, but it is not reset between context, challenge, recovery, or memory-question stages. In particular, the memory questions are asked after the interaction challenges in the same conversational state. The default lifecycle is therefore
\[
\textsc{Reset}
\rightarrow \textsc{Context}
\rightarrow \textsc{Challenge/Recovery}
\rightarrow \textsc{SameSessionQA}
\rightarrow \textsc{Finalize}.
\]
No reset is performed between individual memory questions.

\subsection{Compilation of Ledger Events into Execution Stages}

The event-level representation is compiled into an ordered stage manifest before model inference. The manifest contains event identifiers, source sample ranges, speaker and addressee labels, dialogue acts, semantic turn boundaries, response requirements, and the expected memory operation. The scheduler executes only this manifest; it does not infer the event ordering from a concatenated waveform at runtime.

\begin{table}[t]
\centering
\small
\caption{Execution stages generated from annotated ledger events.}
\label{tab:execution_stages}

\setlength{\tabcolsep}{4pt}
\renewcommand{\arraystretch}{1.15}

\begin{tabularx}{\textwidth}{
    @{}
    >{\raggedright\arraybackslash}p{0.23\textwidth}
    >{\raggedright\arraybackslash}p{0.23\textwidth}
    >{\raggedright\arraybackslash}X
    @{}
}
\toprule
\textbf{Stage} & \textbf{Purpose} & \textbf{Execution semantics} \\
\midrule

\texttt{play\_audio}
&
Context delivery
&
Plays the bounded audio window associated with a user event and preserves the event's speaker, addressee, and dialogue-act metadata.
\\[2pt]

\shortstack[l]{\texttt{wait\_for\_assistant\_}\\\texttt{state}}
&
State-triggered synchronization
&
Waits for decoded assistant speech to satisfy a target state and a minimum active-speech duration. The stage terminates on satisfaction or timeout.
\\[2pt]

\texttt{inject\_interruption}
&
Interruption challenge
&
Delivers the interruption audio only after the preceding speaking-state trigger has been satisfied.
\\[2pt]

\texttt{inject\_backchannel}
&
Backchannel challenge
&
Delivers a backchannel only after the assistant has produced the required amount of active speech.
\\[2pt]

\texttt{play\_side\_talk}
&
Non-addressee speech
&
Plays speech addressed to another participant and then observes whether the assistant responds during the observation interval.
\\[2pt]

\texttt{wait}
&
Annotated pause or observation
&
Advances the scheduler for a fixed duration without fabricating user speech. A pause inside a turn is not treated as a semantic turn boundary.
\\[2pt]

\texttt{end\_user\_event}
&
Event boundary
&
Records the semantic turn boundary and whether an assistant response is required; it does not reset the session.
\\[2pt]

\shortstack[l]{\texttt{wait\_for\_assistant\_}\\\texttt{silence}}
&
Recovery synchronization
&
Waits for stable assistant silence after a response-required event.
\\[2pt]

\shortstack[l]{\texttt{run\_same\_session\_}\\\texttt{memory\_qa}}
&
Memory evaluation
&
Opens the dynamically timed question-and-answer windows without clearing the preceding conversational state.
\\

\bottomrule
\end{tabularx}
\end{table}

For ordinary response-required context events, the manifest inserts a response-start wait followed by a response-silence wait before the next context event. Adaptive interruption and backchannel events instead insert a state-trigger wait immediately before the challenge audio. This ordering ensures that the challenge is aligned to the model's actual speaking state rather than to a fixed timestamp in the source recording.

\subsection{Audio Windows and Clock Semantics}

For each speech event, the scheduler stores an actual source range \([s_i,e_i)\) in samples. The range is derived from the annotated speech window and expanded by a fixed padding of \(120\,\mathrm{ms}\), subject to the available source boundaries. The original waveform is never modified in place. A concatenated waveform may be retained as a compatibility artifact, but it is not the authority for adaptive execution; the stage manifest and its sample ranges are authoritative.

The protocol maintains two clocks. \emph{Media time} measures the amount of user and assistant audio advanced by the scheduler, whereas \emph{wall time} is measured with a monotonic runtime clock. Adaptive sessions are therefore allowed to have different wall-clock durations for different models. Runtime records preserve both clocks, the decoded assistant VAD state, and the timestamps reported by the native audio bridge. Decisions about whether the assistant is speaking are made from decoded PCM audio rather than packet arrival. Input audio is processed in \(80\,\mathrm{ms}\) chunks, and the scheduler polls state at \(20\,\mathrm{ms}\) intervals.

\subsection{Adaptive Challenge Triggering}

Interruption and backchannel challenges use a closed-loop trigger:
\[
\text{wait for assistant speech}
\;\rightarrow\;
\text{verify VAD duration}
\;\rightarrow\;
\text{inject challenge audio}.
\]

An interruption requires at least \(500\,\mathrm{ms}\) of continuous decoded assistant speech, while a backchannel requires at least \(800\,\mathrm{ms}\). Both trigger waits have a \(10\,\mathrm{s}\) assistant-start timeout. If the condition is not satisfied before the timeout, the challenge audio is not sent and the event is recorded as an unrealized challenge with a trigger timeout. In particular, codec activity without VAD-confirmed speech does not satisfy the trigger.

The remaining challenge types use different execution semantics. A pause is implemented as a scheduler wait over its annotated duration and does not force the assistant to speak. 
Speech addressed to another participant is played as audio and
followed by a \(1.2\,\mathrm{s}\) observation interval. Its structured addressee
label is retained in the evaluator's event record; whether an
auxiliary module also accesses this field is a separate input
configuration described in Appendix E. Thus, these events test pause intrusion and target-speaker selectivity, respectively, rather than using the same speech-trigger rule as interruption and backchannel events.

After an event whose semantic turn has ended and which requires a response, the scheduler waits for at least \(160\,\mathrm{ms}\) of assistant speech. It then waits for stable silence for up to \(12\,\mathrm{s}\), with a \(2\,\mathrm{s}\) silence criterion. Failure to observe response onset is recorded as a response-initiation failure; it is not converted into a zero-duration response.

\subsection{Recovery and Stop-Latency Measurement}

For an interruption, the assistant's decoded waveform is monitored from the time at which challenge audio is injected. The runtime record stores whether the assistant was speaking before injection, whether speech continued after injection, the first observed speech offset, and whether the response ended within the configured silence window. Stop latency is computed only when a valid assistant speech interval and a valid interruption injection are both available. If the assistant does not stop before the response timeout, the event remains in the record with an explicit timeout or failure flag.

Backchannels are not treated as interruptions: the protocol records whether the assistant continues, pauses, or restarts after the feedback audio. For non-addressee speech, the observation interval is retained even when the assistant remains silent, since silence is a valid selectivity outcome. All runtime failures are recorded under a ``record-and-continue'' policy whenever the session remains executable. A server crash, unsupported cancellation, or unrecoverable protocol violation terminates the sample and preserves the partial runtime artifacts.

\subsection{Same-Session Memory Questions}

The final stage evaluates memory without resetting the model. Before each question, the controller waits for \(1.2\,\mathrm{s}\) of stable assistant silence, with a maximum pre-input drainage timeout of \(60\,\mathrm{s}\). When supported by the model interface, an outstanding response is cancelled through the native cancellation mechanism without clearing conversational state; the controller then allows an additional \(10\,\mathrm{s}\) recovery interval before sending the question.

For each question, the execution order is
\[
\text{stable-silence check}
\rightarrow
\text{question playback}
\rightarrow
\text{answer-onset wait}
\rightarrow
\text{answer collection}
\rightarrow
\text{silence/turn-end closure}.
\]

The answer-onset timeout and answer-collection timeout are both \(30\,\mathrm{s}\). An answer is closed by a detected turn end, a \(2\,\mathrm{s}\) silence interval, or the answer timeout. Late speech after a timeout is held in a \(10\,\mathrm{s}\) quarantine interval and is not retroactively assigned to the expired question. If no answer is observed, the question is marked \texttt{no\_response}; text from a previous question is never reused.

Memory questions and reference targets are fixed by the planned
session. If a speech-triggered challenge is not delivered, these
targets are not reconstructed from the realized playback history.
Fixed-denominator memory scores therefore measure end-to-end task
completion and can reflect both missed evidence delivery and
subsequent memory or response errors. They do not isolate retention
conditional on receiving the supporting evidence.

\subsection{Validity Gates and Failure Accounting}

The protocol reports validity at multiple levels. First, a session is \emph{protocol-valid} only if its manifest, audio windows, runtime clocks, and required artifacts pass structural validation. Second, a valid session is \emph{behaviorally observable} only if decoded assistant audio is available for the relevant decision windows. Third, each planned challenge is labeled as realized, unrealized, or timed out according to its trigger result. These states are not collapsed:
\[
\begin{array}{c}
\text{attempted session}
\\[-1mm]
\quad\downarrow
\\[-1mm]
\text{protocol valid}
\;\;\text{vs.}\;\;
\text{protocol invalid}
\\[-1mm]
\quad\downarrow
\\[-1mm]
\text{behaviorally observable}
\;\;\text{vs.}\;\;
\text{not observable}
\\[-1mm]
\quad\downarrow
\\[-1mm]
\text{challenge realized}
\;\;\text{vs.}\;\;
\text{unrealized/timeout}.
\end{array}
\]

Consequently, conditional behavioral metrics are computed only on realized and observable challenges, while challenge-realization and trigger-timeout rates are reported with the planned-challenge denominator. Runtime failures, response-initiation failures, no-response questions, skipped challenge audio, and timeout events remain in the released execution records rather than being silently discarded.

\subsection{Reproducible Protocol Configuration}

\begin{table}[t]
\centering
\small
\caption{Default timing and execution parameters.}
\label{tab:execution_configuration}
\begin{tabular}{ll@{\qquad}ll}
\toprule
Parameter & Value & Parameter & Value \\
\midrule
Input chunk & \(80\,\mathrm{ms}\) & Scheduler poll & \(20\,\mathrm{ms}\) \\
Event padding & \(120\,\mathrm{ms}\) & Assistant-start timeout & \(10\,\mathrm{s}\) \\
Interruption trigger speech & \(500\,\mathrm{ms}\) & Backchannel trigger speech & \(800\,\mathrm{ms}\) \\
Minimum response speech & \(160\,\mathrm{ms}\) & Assistant-end timeout & \(12\,\mathrm{s}\) \\
Post-event silence & \(240\,\mathrm{ms}\) & Stable assistant silence & \(2\,\mathrm{s}\) \\
Side-talk observation & \(1.2\,\mathrm{s}\) & QA pre-input quiet & \(1.2\,\mathrm{s}\) \\
QA start timeout & \(30\,\mathrm{s}\) & QA answer timeout & \(30\,\mathrm{s}\) \\
QA end silence & \(2\,\mathrm{s}\) & Late-response quarantine & \(10\,\mathrm{s}\) \\
Maximum session duration & \(900\,\mathrm{s}\) & Maximum input stream RTF & \(1.25\) \\
\bottomrule
\end{tabular}
\end{table}

All parameters in Table~\ref{tab:execution_configuration}, including VAD thresholds, trigger conditions, timeout behavior, and session-reset rules, are frozen before evaluation. The runtime output records the effective configuration, stage identifiers, media and wall-clock timestamps, VAD decisions, trigger outcomes, timeout flags, and failure types, thereby making each adaptive execution auditable and reproducible.

\section{Metric Definitions and Denominators}
\label{app:metrics}

This section defines the aggregate metrics used to evaluate full-duplex interaction and same-session memory behavior. Every metric is first computed independently for each sample and is then aggregated across samples using a macro-average. The resulting values therefore give every valid sample equal weight, regardless of the number of events or the duration of its interaction.

\subsection{Aggregation and Missing-Value Policy}

Let \(m_i\) denote the value of metric \(m\) for sample \(i\), and let \(\mathcal{I}_m\) be the set of samples for which \(m_i\) is defined. The reported aggregate is
\begin{equation}
\overline{m}
=
\frac{1}{|\mathcal{I}_m|}
\sum_{i\in\mathcal{I}_m} m_i .
\label{eq:metric_macro_average}
\end{equation}
If \(\mathcal{I}_m\) is empty, the metric is reported as unavailable. Missing values are not replaced with zero. An unanswered memory question is not a missing observation:
it remains in the fixed question denominator and receives no credit.
By contrast, an interaction metric without eligible observations
is undefined and is excluded from that metric's mean. An unrealized
challenge remains in the planned-challenge denominator for RCR
but does not supply an observation for delivery-conditioned metrics.

All ratios are computed using sample-specific denominators before aggregation. The evaluation therefore does not pool all events, questions, or audio segments across the complete test set. This rule also applies to latency statistics: the median and the 90th percentile are first computed within each sample and are then averaged across samples. Thus, \texttt{stop\_latency\_p50\_s} and \texttt{stop\_latency\_p90\_s} are macro-averaged sample-level quantiles, rather than quantiles computed from one pooled latency distribution.

\subsection{Runtime Validity and Behavioral Observability}

Protocol validity and behavioral observability are reported separately. For sample \(i\), protocol validity is defined as
\begin{equation}
v_i =
\mathbf{1}\!\left[
\text{runtime timestamps are valid}
\land
\text{no runtime failure is recorded}
\land
\text{a decoded assistant-audio artifact exists}
\right].
\end{equation}
The aggregate \texttt{protocol\_valid} is the mean of \(v_i\).

Behavioral observability is defined as
\begin{equation}
o_i =
\mathbf{1}\!\left[
d_i^{\mathrm{asst}}>0
\land
n_i^{\mathrm{raw}}>1
\right],
\end{equation}
where \(d_i^{\mathrm{asst}}\) is the decoded assistant-speech duration and \(n_i^{\mathrm{raw}}\) is the number of samples in the raw assistant track. The aggregate \texttt{behaviorally\_observable} is the mean of \(o_i\). Consequently, a session can be protocol-valid while remaining behaviorally unobservable if the model produces no usable decoded speech.

\subsection{Adaptive-Challenge Metrics}

Let \(\mathcal{P}_i\) be the set of distinct challenge identifiers planned in the manifest of sample \(i\), and let \(\mathcal{R}_i\) be the subset recorded as realized in the runtime trace. For an interruption or backchannel, realization requires successful audio injection after the corresponding assistant-speaking trigger has been satisfied. For wait-based challenges, realization follows the corresponding runtime observation rule. The sample-level challenge realization rate is
\begin{equation}
r_i =
\frac{|\mathcal{R}_i|}{|\mathcal{P}_i|}.
\end{equation}
Samples without a planned challenge do not contribute to this ratio.

Let \(T_i\) denote the number of challenge-trigger waits that time out before the required assistant state is observed. The trigger-timeout rate is
\begin{equation}
t_i =
\frac{T_i}{|\mathcal{P}_i|}.
\end{equation}
When a trigger times out, the corresponding interruption or backchannel audio is not injected and the challenge is not counted as realized.

The conditional full-duplex score, denoted by \(c_i\), summarizes behavioral correctness on challenge instances for which the corresponding scenario can be evaluated. The scenario-specific correctness criteria are:

\begin{itemize}
    \item timely stopping after an interruption;
    \item continuation of the ongoing response after a backchannel;
    \item absence of assistant intrusion during an annotated pause; and
    \item target-speaker selectivity during speech addressed to another participant.
\end{itemize}

The aggregate conditional score is
\begin{equation}
\texttt{conditional\_cfd}
=
\frac{1}{|\mathcal{I}_{c}|}
\sum_{i\in\mathcal{I}_{c}} c_i,
\end{equation}
where \(\mathcal{I}_{c}\) contains samples with a defined conditional score. Because this score is conditioned on behaviorally scorable challenges, it must be interpreted together with \texttt{challenge\_realization\_rate}, \texttt{trigger\_timeout\_rate}, and \texttt{behaviorally\_observable}.

The aggregate \texttt{adaptive\_fd} is the macro-average of the corresponding sample-level adaptive full-duplex scores. It is an auxiliary summary that combines challenge opportunity and challenge behavior. It is not recomputed as the product of separately averaged aggregate quantities.

\subsection{Ordinary-Turn Timing Metrics}

Let \(\mathcal{U}_i\) be the set of response-required semantic turns in sample \(i\), and let \(\mathcal{O}_i\subseteq\mathcal{U}_i\) be the subset satisfying the ordinary-turn eligibility conditions. An eligible ordinary turn must have a valid semantic boundary, valid speech-alignment markers, usable native audio-clock metadata, and an input stream within the prescribed real-time limit.

Response coverage is computed over \(\mathcal{U}_i\):
\begin{equation}
\texttt{response\_coverage}_i
=
\frac{
\sum_{u\in\mathcal{U}_i}
\mathbf{1}[\text{assistant speech is observed for }u]
}{
|\mathcal{U}_i|
}.
\end{equation}

The remaining ordinary-turn metrics use \(\mathcal{O}_i\):
\begin{align}
\texttt{premature\_onset\_rate}_i
&=
\frac{
\sum_{u\in\mathcal{O}_i}
\mathbf{1}[\text{assistant onset precedes user end}]
}{
|\mathcal{O}_i|
},
\\
\texttt{carry\_in\_rate}_i
&=
\frac{
\sum_{u\in\mathcal{O}_i}
\mathbf{1}[\text{assistant is speaking at input onset}]
}{
|\mathcal{O}_i|
},
\\
\texttt{overlap\_before\_end\_rate}_i
&=
\frac{
\sum_{u\in\mathcal{O}_i}
\mathbf{1}[\text{assistant overlaps user speech before user end}]
}{
|\mathcal{O}_i|
}.
\end{align}

For pause handling, let \(\mathcal{W}_i\) be the annotated pause-wait events. Pause intrusion is defined as
\begin{equation}
\texttt{pause\_intrusion\_rate}_i
=
\frac{
\sum_{w\in\mathcal{W}_i}
\mathbf{1}[\text{assistant speech is observed during }w]
}{
|\mathcal{W}_i|
}.
\end{equation}

These definitions distinguish a new premature response from speech carried over from a previous turn and use only turns with valid timing evidence.

\subsection{Interruption Stop Metrics}

Let \(\mathcal{L}_i\) be the set of valid stop latencies for realized interruption events in sample \(i\). A stop latency is included only when the interruption was actually injected and the decoded assistant waveform contains a valid speech-offset measurement. The sample-level latency statistics are
\begin{align}
\ell_i^{50}
&=
\operatorname{Quantile}_{0.50}(\mathcal{L}_i),
\\
\ell_i^{90}
&=
\operatorname{Quantile}_{0.90}(\mathcal{L}_i),
\\
s_i^{800}
&=
\frac{
\sum_{\ell\in\mathcal{L}_i}
\mathbf{1}[\ell\leq 0.8\ \mathrm{s}]
}{
|\mathcal{L}_i|
}.
\end{align}

The macro-averages of these quantities are reported as \texttt{stop\_latency\_p50\_s}, \texttt{stop\_latency\_p90\_s}, and \texttt{stop\_success\_at\_800ms}, respectively. Samples without a valid stop-latency observation are excluded from the corresponding aggregate and are not assigned a latency of zero.

The metric \texttt{assistant\_speech\_duration\_s} is the macro-average of the total decoded assistant-speech duration per sample. It is a descriptive quantity rather than a quality measure.

\begin{table*}[t]
\centering
\scriptsize
\caption{Aggregate runtime metrics and their sample-level denominators.}
\label{tab:aggregate_runtime_metrics}
\begin{tabular}{p{0.26\textwidth}c p{0.54\textwidth}}
\toprule
\textbf{Metric} & \textbf{Direction} & \textbf{Denominator or interpretation} \\
\midrule
\texttt{protocol\_valid} & \(\uparrow\) & Attempted samples. \\
\texttt{behaviorally\_observable} & \(\uparrow\) & Attempted samples; requires usable decoded assistant speech. \\
\texttt{challenge\_realization\_rate} & \(\uparrow\) & Distinct planned challenge identifiers. \\
\texttt{conditional\_cfd} & \(\uparrow^{\dagger}\) & Challenge instances eligible for conditional behavioral scoring. \\
\texttt{adaptive\_fd} & \(\uparrow^{\dagger}\) & Sample-level adaptive full-duplex score; auxiliary summary. \\
\texttt{trigger\_timeout\_rate} & \(\downarrow\) & Distinct planned challenges. \\
\texttt{response\_coverage} & \(\uparrow\) & Response-required semantic turns. \\
\texttt{premature\_onset\_rate} & \(\downarrow\) & Eligible ordinary response turns. \\
\texttt{carry\_in\_rate} & \(\downarrow\) & Eligible ordinary response turns. \\
\texttt{overlap\_before\_end\_rate} & \(\downarrow\) & Eligible ordinary response turns. \\
\texttt{pause\_intrusion\_rate} & \(\downarrow\) & Annotated pause-wait events. \\
\texttt{stop\_latency\_p50\_s} & \(\downarrow\) & Valid interruption stop latencies, summarized within each sample. \\
\texttt{stop\_latency\_p90\_s} & \(\downarrow\) & Valid interruption stop latencies, summarized within each sample. \\
\texttt{stop\_success\_at\_800ms} & \(\uparrow\) & Valid interruption stop-latency observations. \\
\texttt{assistant\_speech\_duration\_s} & -- & Total decoded assistant-speech duration; descriptive only. \\
\bottomrule
\end{tabular}

\vspace{1mm}
\raggedright
\scriptsize
\(\dagger\) Conditional or auxiliary measures that must be interpreted with challenge realization, trigger timeout, and observability statistics.
\end{table*}

\subsection{Memory-Answer Normalization}

Memory answers are obtained from the transcript associated with each question-specific answer window. When available, the frozen transcript associated with the answer window is used; otherwise, the corresponding answer artifact is used. Text is normalized with Unicode NFKC normalization, lowercasing, and removal of punctuation and symbols. English text is tokenized by whitespace-delimited words, while Chinese text is tokenized into CJK characters and alphanumeric units.

For token sequences \(A\) and \(B\), multiset precision and recall are
\begin{equation}
P(A,B)=\frac{|A\cap B|}{|B|},
\qquad
R(A,B)=\frac{|A\cap B|}{|A|},
\end{equation}
and the token overlap score is
\begin{equation}
F_1(A,B)
=
\frac{2P(A,B)R(A,B)}{P(A,B)+R(A,B)}.
\end{equation}
A value is considered present when its normalized form is an exact substring of the answer or when token \(F_1\) reaches \(0.80\) for values of at most two tokens and \(0.60\) for longer values.

\subsection{Memory QA Metrics}

Let \(\mathcal{Q}_i\) be all memory questions in sample \(i\), and let \(\mathcal{Q}_i^{+}\) be the subset with a non-empty response that is not marked as \texttt{no\_response}. Response and no-response rates are
\begin{align}
\texttt{response\_rate}_i
&=
\frac{|\mathcal{Q}_i^{+}|}{|\mathcal{Q}_i|},
\\
\texttt{no\_response\_rate}_i
&=
1-\texttt{response\_rate}_i.
\end{align}

Let \(\mathcal{Q}_i^{\mathrm{latest}}\) denote the latest-value questions. For question \(q\), let \(a_q\) indicate that an accepted updated value is present and let \(f_q\) indicate that a forbidden superseded value is present. Then
\begin{align}
\texttt{latest\_value\_accuracy}_i
&=
\frac{
\sum_{q\in\mathcal{Q}_i^{\mathrm{latest}}}
\mathbf{1}[q\text{ answered}\land a_q\land\neg f_q]
}{
|\mathcal{Q}_i^{\mathrm{latest}}|
},
\\
\texttt{superseded\_value\_error}_i
&=
\frac{
\sum_{q\in\mathcal{Q}_i^{\mathrm{latest}}}
\mathbf{1}[q\text{ answered}\land f_q]
}{
|\mathcal{Q}_i^{\mathrm{latest}}|
}.
\end{align}

Accuracy conditioned on producing an answer is
\begin{equation}
\texttt{answer\_accuracy\_given\_response}_i
=
\frac{
\sum_{q\in\mathcal{Q}_i^{\mathrm{latest}}\cap\mathcal{Q}_i^{+}}
\mathbf{1}[a_q\land\neg f_q]
}{
|\mathcal{Q}_i^{\mathrm{latest}}\cap\mathcal{Q}_i^{+}|
}.
\end{equation}
This metric separates incorrect answers from failures to answer.

Let \(\mathcal{Q}_i^{\mathrm{dist}}\) denote the distractor-ignore questions. A distractor is considered adopted when the answer is non-empty and does not explicitly reject the distractor, or when the answer directly matches distractor content without rejection. A distractor is considered rejected only when the model responds and the adoption condition is false:
\begin{align}
\texttt{distractor\_adoption\_rate}_i
&=
\frac{
\sum_{q\in\mathcal{Q}_i^{\mathrm{dist}}}
\mathbf{1}[\text{distractor adopted}]
}{
|\mathcal{Q}_i^{\mathrm{dist}}|
},
\\
\texttt{distractor\_rejection\_accuracy}_i
&=
\frac{
\sum_{q\in\mathcal{Q}_i^{\mathrm{dist}}}
\mathbf{1}[q\text{ answered}\land\text{distractor not adopted}]
}{
|\mathcal{Q}_i^{\mathrm{dist}}|
}.
\end{align}
These two measures are reported together with \texttt{response\_rate}, since silence can reduce distractor adoption without demonstrating correct rejection.

For a final-summary question, the reference summary is split into atomic facts using sentence-final punctuation, semicolons, question marks, and line breaks. A fact is considered covered when its token \(F_1\) with the answer is at least \(0.50\). If \(\mathcal{F}_q\) is the reference fact set for question \(q\), then
\begin{equation}
\operatorname{coverage}(q)
=
\frac{
\sum_{f\in\mathcal{F}_q}
\mathbf{1}[F_1(f,\widehat{a}_q)\geq 0.50]
}{
|\mathcal{F}_q|
}.
\end{equation}
The sample-level \texttt{summary\_fact\_coverage} is the mean coverage over final-summary questions.

The metric \texttt{legacy\_token\_f1} is the mean token \(F_1\) between each answer and its full reference answer. It is retained for compatibility with earlier result tables, but it is not used as the sole measure of memory quality.

\begin{table*}[t]
\centering
\scriptsize
\caption{Aggregate same-session memory metrics and their sample-level denominators.}
\label{tab:aggregate_memory_metrics}
\begin{tabular}{p{0.29\textwidth}c p{0.52\textwidth}}
\toprule
\textbf{Metric} & \textbf{Direction} & \textbf{Denominator or interpretation} \\
\midrule
\texttt{response\_rate} & \(\uparrow\) & All memory questions. \\
\texttt{no\_response\_rate} & \(\downarrow\) & All memory questions; complementary to response rate. \\
\texttt{latest\_value\_accuracy} & \(\uparrow\) & All latest-value questions, including unanswered questions. \\
\texttt{superseded\_value\_error} & \(\downarrow\) & All latest-value questions. \\
\texttt{distractor\_adoption\_rate} & \(\downarrow\) & All distractor-ignore questions. \\
\texttt{distractor\_rejection\_accuracy} & \(\uparrow\) & All distractor-ignore questions; rejection additionally requires a response. \\
\texttt{summary\_fact\_coverage} & \(\uparrow\) & Atomic reference facts in final-summary questions. \\
\texttt{answer\_accuracy\_given\_response} & \(\uparrow\) & Answered latest-value questions only. \\
\texttt{legacy\_token\_f1} & \(\uparrow\) & All memory questions; retained as a compatibility measure. \\
\bottomrule
\end{tabular}
\end{table*}

\subsection{Denominator Reporting and Interpretation}

For each evaluated model, the results report the total number of attempted samples, the number of protocol-valid samples, and the number of behaviorally observable samples. The total number of memory questions and the number of latest-value questions are also reported. Challenge, ordinary-turn, pause, and stop-latency denominators are retained in the sample-level records used for aggregation.

Conditional metrics must always be reported together with their coverage variables. In particular, \texttt{conditional\_cfd} is accompanied by challenge realization and trigger-timeout rates; stop-latency measures are accompanied by the number of valid latency observations; and conditional memory accuracy is accompanied by response rate. A low error rate from a nearly silent model is therefore not interpreted as evidence of strong full-duplex or memory behavior. Stratified analyses by scenario or language reuse the definitions above and do not introduce additional metrics.

\section{Baseline Models and Inference Configurations}
\label{app:baseline_details}

\subsection{Evaluation Scope and Comparison Protocol}

We evaluate five publicly available full-duplex speech models: \texttt{MiniCPM}-o~4.5~\citep{minicpmo45}, \texttt{Freeze-Omni}~\citep{freezeomni}, \texttt{Moshi}~\citep{moshi}, \texttt{Lychee-FD}~\citep{lycheefd}, and \texttt{PersonaPlex}-7B-v1~\citep{personaplex}. The models are selected to cover different streaming speech architectures and native interaction interfaces. All baseline models use the same session manifests, planned event
order, source audio clips, and scheduling rules. Because
speech-triggered challenges depend on each model's realized output,
the subset of challenge clips actually delivered can differ
across models. A sample is executed as one persistent session: the session is reset before the sample begins, but not between context events, full-duplex challenges, recovery stages, or memory questions.

The baseline speech models are not provided with gold memory states,
reference answers, alignment boundaries, event labels, addressee
labels, or future events. This restriction concerns baseline-model
inputs; the interfaces available to ACT-Mem's auxiliary modules
are specified separately in Appendix E. All models are evaluated with the same adaptive execution controller, the same challenge-triggering rules, the same session-reset policy, and the same answer-attribution procedure. Model-specific native chunk sizes, sampling rates, voice prompts, and transport protocols are retained when they are required by the corresponding implementation. These interface-specific parameters are reported explicitly rather than hidden under a common configuration.

\subsection{Native Interfaces and Model Configurations}

Table~\ref{tab:baseline_model_configuration} summarizes the model-specific settings used for inference. Input and output sample rates refer to the native model bridge; the evaluator maintains a shared reference clock and performs any required resampling. A native inference window is distinct from the evaluator's event window and does not change the logical event sequence.

\begin{table*}[t]
\centering
\scriptsize
\caption{Native model interfaces and model-specific inference settings.}
\label{tab:baseline_model_configuration}
\begin{tabular}{p{0.13\textwidth}p{0.15\textwidth}p{0.1\textwidth}p{0.16\textwidth}p{0.18\textwidth}c}
\toprule
\textbf{Model} & \textbf{Native interface} & \textbf{Input / output} & \textbf{Chunk or inference window} & \textbf{Model-specific condition} & \textbf{GPU} \\
\midrule
\texttt{MiniCPM}-o~4.5 & Native streaming worker & \(16/24\) kHz & \(80\) ms input; \(1000\) ms native window & Greedy decoding, fixed seed, fixed reference audio & 1 \\
\texttt{Freeze-Omni} & Socket.IO-based streaming service & \(16/24\) kHz & \(20\) ms native input block & \(p=0\), \(k=1\), temperature \(=1.0\); neutral system prompt & 1 \\
\texttt{Moshi} & WebSocket streaming service & \(24/24\) kHz & \(80\) ms input block & Native server decoding; no evaluator-side task-specific sampling override & 1 \\
\texttt{Lychee-FD} & vLLM-backed streaming service & \(16/24\) kHz & \(80\) ms input; \(400\) ms native inference window & Fixed prompt voice; native backend decoding & 2 \\
\texttt{PersonaPlex}-7B-v1 & Native WebSocket streaming service & \(24/24\) kHz & \(80\) ms input block & Fixed voice prompt and fixed random seed & 1 \\
\bottomrule
\end{tabular}
\end{table*}

\texttt{MiniCPM}-o~4.5 uses greedy decoding with the frozen evaluation seed and a fixed reference audio prompt. \texttt{Freeze-Omni} uses the native service configuration with \(p=0\), \(k=1\), and temperature \(1.0\). \texttt{Moshi} is evaluated through its native streaming service without introducing an additional task-specific decoding rule. \texttt{Lychee-FD} uses the vLLM-backed native inference path with a fixed prompt voice and a \(400\,\mathrm{ms}\) native inference window. \texttt{PersonaPlex} uses a fixed voice prompt and the evaluation seed for all sessions.

The native interfaces do not expose identical user-turn controls. \texttt{Freeze-Omni} supports an explicit response-boundary operation, whereas the other bridges record local event boundaries when an equivalent native message is unavailable. These local markers advance the evaluator's state machine but do not reset the model or claim that the model received an explicit semantic end-of-turn signal. This distinction is retained in the runtime artifacts.

\subsection{Shared Evaluator-Side Configuration}

All models are evaluated with a common evaluator-side configuration. User audio is streamed in real time as bounded event clips. Each delivered clip is played once, is not accelerated, and is not
truncated because of the assistant's current output. Challenges
whose trigger conditions are not met are omitted as specified below. The controller uses an \(80\,\mathrm{ms}\) reference input chunk, a \(20\,\mathrm{ms}\) polling interval, and a maximum input real-time factor of \(1.25\). Native model sample rates and chunk sizes are permitted, but all timing decisions are mapped to the shared reference clock.

Assistant speaking state is determined from decoded assistant PCM using the shared waveform-level VAD. Packet arrival, codec activity, and text-token activity are not treated as evidence of speech. Interruption audio is injected only after at least \(500\,\mathrm{ms}\) of continuous decoded assistant speech, and backchannel audio is injected only after at least \(800\,\mathrm{ms}\) of such speech. If the trigger condition is not satisfied within the \(10\,\mathrm{s}\) trigger window, the challenge is recorded as unrealized and its audio is not injected.

The common protocol parameters are listed in Table~\ref{tab:baseline_shared_configuration}. These parameters are frozen before evaluation and are not tuned on held-out sessions.

\begin{table}[t]
\centering
\scriptsize
\caption{Shared evaluator-side inference and timing configuration.}
\label{tab:baseline_shared_configuration}
\begin{tabular}{p{0.53\linewidth}l}
\toprule
\textbf{Parameter} & \textbf{Value} \\
\midrule
Reference input chunk & \(80\,\mathrm{ms}\) \\
Evaluator polling interval & \(20\,\mathrm{ms}\) \\
Event-window padding & \(120\,\mathrm{ms}\) \\
Assistant-start timeout & \(10\,\mathrm{s}\) \\
Assistant-end timeout & \(12\,\mathrm{s}\) \\
Interruption trigger speech & \(500\,\mathrm{ms}\) \\
Backchannel trigger speech & \(800\,\mathrm{ms}\) \\
Minimum response speech & \(160\,\mathrm{ms}\) \\
Stable assistant silence & \(2\,\mathrm{s}\) \\
Side-talk observation window & \(1.2\,\mathrm{s}\) \\
QA start timeout & \(30\,\mathrm{s}\) \\
QA answer timeout & \(30\,\mathrm{s}\) \\
QA pre-input quiet period & \(1.2\,\mathrm{s}\) \\
QA pre-input drainage timeout & \(60\,\mathrm{s}\) \\
Late-response quarantine & \(10\,\mathrm{s}\) \\
Maximum session duration & \(900\,\mathrm{s}\) \\
\bottomrule
\end{tabular}
\end{table}

\subsection{Prompt and Information Isolation}

All baseline models receive a neutral voice-assistant instruction
in the language of the corresponding sample. The default English instruction is ``You are a voice assistant,'' with the corresponding neutral Chinese instruction for Chinese samples. The instruction is transported through the native interface when required by the model, but no task-specific memory or interaction instruction is added.

The following information is not exposed to any baseline model:

\begin{itemize}
    \item gold ledger states or future memory updates;
    \item event types, challenge identifiers, or addressee labels;
    \item MFA word boundaries or evaluator-side VAD decisions;
    \item reference answers and expected memory values; and
    \item assistant reference audio generated from the benchmark annotations.
\end{itemize}

A model may therefore use only its native audio history and the neutral instruction supplied at session initialization. The proposed memory module, when evaluated in the controlled comparison, uses the same frozen \texttt{MiniCPM}-o~4.5 backbone and the same audio protocol; its additional memory conditioning is applied only through the method-specific inference path described in the main paper and the implementation appendix.

\subsection{Audio Output, Transcription, and Answer Attribution}

All assistant audio is captured from the native model bridge and retained as a decoded waveform. The waveform is used for VAD-based state detection, challenge triggering, stop-latency measurement, and response-onset attribution. Memory answers are evaluated from independent answer windows rather than from a concatenated session transcript.

For cross-model textual comparison, each answer waveform is transcribed with the same frozen external Whisper-small configuration. The language is taken from the sample manifest, decoding temperature is set to zero, and the same transcription backend is used for all models. Native model text, when available, is retained as a secondary artifact but is not used as the primary cross-model transcript. A transcription failure is recorded explicitly and makes the corresponding primary textual score unavailable; it is not silently converted to a zero-valued answer.

The answer window opens after decoded assistant speech onset and closes on decoded silence, an explicit turn boundary, or the configured answer timeout. A late response after timeout is kept in a quarantine interval and is not assigned retroactively to the expired question. No answer text is reused between questions.

\subsection{Calibration, Test Isolation, and Reproducibility}

Before the held-out evaluation, a fixed calibration split is used to freeze the shared VAD, challenge-trigger thresholds, timeout values, text normalization rules, and scoring parameters. The held-out sessions are not used for prompt tuning, decoding-parameter tuning, VAD adjustment, timeout selection, or checkpoint selection.

For every model, the evaluation record stores the public model identifier, checkpoint identity, repository revision when available, model-profile configuration, prompt transport, decoding settings, random seed, native sample-rate configuration, assigned GPU resources, and runtime environment identity. Checkpoint and configuration hashes are retained in the artifact manifest so that the reported results can be traced to an unambiguous model state without exposing machine-specific filesystem paths in the paper.

GPU isolation is enforced during inference. A run is considered invalid for paper reporting if an unrelated process contaminates an assigned GPU. The same session and sample identifiers are preserved across all runtime artifacts, allowing failed samples and resumed samples to be audited without double counting.

\subsection{Failure Handling and Effective Denominators}

Runtime failures, server disconnects, unsupported native controls, trigger timeouts, answer timeouts, and no-response questions are retained in the execution record. A failed sample is not silently removed from the attempted sample count. Instead, the evaluation reports protocol validity, behavioral observability, completion or failure status, and the effective denominator of every metric.

A challenge that is not realized because its trigger condition is not reached is distinguished from a challenge that is realized but handled incorrectly. Likewise, a valid session with no assistant speech is distinguished from a protocol-invalid session. Conditional behavioral scores are computed only when their required audio and timing evidence exists, while challenge realization, trigger timeout, and failure counts are reported separately. This accounting prevents differences in interface robustness, speech observability, and challenge exposure from being mistaken for differences in full-duplex behavior or memory fidelity.

\section{ACT-Mem Implementation Details}
\label{app:actmem_implementation}

\subsection{Implementation Scope and Frozen Components}

\textsf{ACT-Mem} is implemented as an external memory module coupled to a frozen \texttt{MiniCPM}-o~4.5 full-duplex backbone~\citep{minicpmo45}. The backbone, including its acoustic encoder, streaming decoder, voice-generation components, and native full-duplex session manager, remains frozen throughout training and evaluation. A frozen speech transcription and timing component provides event-aligned textual observations to the memory controller. The trainable part consists of an event controller and a typed retriever, which jointly contain \(8.92\)M parameters.

The online computation is organized into three stages:
\begin{equation}
\text{event observation}
\;\longrightarrow\;
\text{transactional memory update}
\;\longrightarrow\;
\text{typed QA retrieval}.
\label{eq:actmem_pipeline}
\end{equation}
Context and full-duplex challenge events update the external memory state. During memory QA, the retriever constructs a compact evidence card from the active state, and the card is provided to the frozen backbone through the native text-prefill interface. The full-duplex audio stream and the memory controller therefore share the same persistent session.

\subsection{Event Controller}

For each delivered event $e_t$, a frozen ASR component transcribes
the evaluator-delimited audio segment. The scheduler supplies the
event index and speaker/addressee metadata. These metadata are
distinct from the controller's learned predictions: a
non-user or non-addressed event can veto admission through an
explicit gate. The adapter uses the default audibility flag rather
than supplying a separate online audibility estimate.

A lightweight text encoder and recurrent state module process the
transcript and chronological context. Four prediction heads produce
the operation, addressee, semantic-completeness, and confidence
estimates:
\begin{equation}
\left(
\widehat{o}_t,
\widehat{a}_t,
\widehat{c}_t,
p_t^{\mathrm{conf}}
\right)
=
f_{\theta}
\left(
x_t,h_{t-1}
\right),
\label{eq:actmem_controller}
\end{equation}
where
\[
\widehat{o}_t\in
\{\textsc{Write},\textsc{Update},\textsc{Ignore},\textsc{Retain}\},
\qquad
\widehat{a}_t\in
\{\text{assistant},\text{other}\},
\qquad
\widehat{c}_t\in\{0,1\}.
\]

The controller proposes atomic entity--value slots from each event. Each slot stores an entity key, value, event identifier, chronological event index, operation type, confidence, source role, and semantic-completeness flag. The controller output is converted into an admission score
\begin{equation}
g_t =
p_t^{\mathrm{addr}}
p_t^{\mathrm{complete}}
\left[
p_t(\textsc{Write})+
p_t(\textsc{Update})
\right]
c_t,
\label{eq:actmem_admission_appendix}
\end{equation}
where \(p_t^{\mathrm{addr}}\), \(p_t^{\mathrm{complete}}\), and \(c_t\) denote the predicted addressee probability, completeness probability, and controller confidence, respectively.

A proposal is processed through the following transaction states:
\[
\textsc{Prepare}
\rightarrow
\begin{cases}
\textsc{Commit}, & g_t\geq \tau_{\mathrm{write}},\\
\textsc{Abort}, & \widehat{a}_t=\text{other}\ \text{or the event is inadmissible},\\
\textsc{Uncertain}, & \text{otherwise}.
\end{cases}
\]
The calibration split selects \(\tau_{\mathrm{write}}=0.35\). Committed proposals are passed to the versioned memory store, while aborted and uncertain proposals remain unavailable to ordinary retrieval.

\subsection{Versioned Transactional Ledger}

The memory store maintains active values, superseded versions, uncertain proposals, and provenance metadata. Each committed item is represented as
\[
m_j=
(k_j,v_j,\tau_j,\iota_j,\rho_j,\gamma_j,\sigma_j),
\]
where \(k_j\) is the entity key, \(v_j\) is the value, \(\tau_j\) is the source-event index, \(\iota_j\) is the source event identifier, \(\rho_j\) is the operation label, \(\gamma_j\) is the confidence, and \(\sigma_j\in\{\textit{active},\textit{superseded},\textit{uncertain}\}\) is the state status.

A \textsc{Write} operation creates a new active item. An \textsc{Update} operation creates a new active version and supersedes the previous active value associated with the same entity key:
\begin{equation}
k_j=k_i,\qquad
\tau_j>\tau_i,\qquad
\rho_j=\textsc{Update}.
\label{eq:actmem_version_condition}
\end{equation}
The predecessor is moved to the superseded store and linked through \(\texttt{supersedes}\) and \(\texttt{superseded\_by}\) fields. A \textsc{Write} or \textsc{Update} proposal whose admission score is below the threshold is retained as uncertain. \textsc{Ignore} records an aborted proposal, whereas \textsc{Retain} preserves the currently active state.

The ledger has a maximum active capacity of 32 slots. When the capacity is exceeded, lower-confidence and older active items are moved to the uncertain state while supersession links remain available for audit. Retrieval uses active items by default; superseded and uncertain items are retained for diagnostic analysis and error attribution.

\subsection{Typed Retrieval and Evidence Cards}

The adapter supplies a query type to the retriever. In the provided
implementation, an annotated type is loaded from the configured
QA metadata when available; otherwise, a deterministic rule routes
the ASR-transcribed question. These annotation-assisted and
text-routed modes must be distinguished when interpreting results.
The query-type set is:
\[
\mathcal{K}
=
\{
\texttt{latest\_value\_qa},
\texttt{ordinary\_memory\_qa},
\texttt{final\_summary\_qa},
\texttt{distractor\_ignore}
\}.
\]
The retrieval scope and maximum number of returned items depend on this type.

\begin{table}[t]
\centering
\scriptsize
\caption{Question routing and retrieval scope.}
\label{tab:actmem_retrieval_scope}
\begin{tabular}{p{0.18\linewidth}p{0.49\linewidth}c}
\toprule
\textbf{Question type} & \textbf{Retrieval behavior} & \textbf{Maximum items} \\
\midrule
Latest-value QA & Selects the target entity's current active value after ranking candidate facts. & 2 candidates, 1 final value \\
Ordinary memory QA & Retrieves active facts relevant to the question. & 4 \\
Final-summary QA & Aggregates active facts across entities and memory types. & 8 \\
Distractor-ignore QA & Routes directly to rejection behavior without memory retrieval. & 0 \\
\bottomrule
\end{tabular}
\end{table}

For a question \(q\) and memory item \(m\), the retriever combines representation similarity, lexical overlap, recency, confidence, and state-based penalties:
\begin{equation}
S(q,m)
=
S_{\mathrm{dense}}(q,m)
+\alpha_{\mathrm{lex}}S_{\mathrm{lex}}(q,m)
+\alpha_{\mathrm{conf}}\gamma_m
+\alpha_{\mathrm{rec}}S_{\mathrm{rec}}(q,m)
-\alpha_{\mathrm{bad}}B(m),
\label{eq:actmem_retrieval_score}
\end{equation}
where \(B(m)\) penalizes superseded, uncertain, or distractor items. The retriever ranks candidate items by \(S(q,m)\) and applies a score-and-margin gate. An evidence card is generated only when the retrieved evidence passes the gate.

The evidence card contains the question type, entity keys, active values, source event identifiers, and a compact answering instruction:
\begin{equation}
C_q =
\operatorname{Serialize}
\left(
q,\,
\{m_j:\,m_j\in\mathcal{R}(q)\}
\right).
\label{eq:actmem_evidence_card}
\end{equation}
The card is provided once at the corresponding QA boundary through the backbone's native text-prefill interface. Context events and full-duplex challenge events continue through the native audio path, while memory conditioning is applied only to the associated same-session question.

The evidence-card intervention begins at the corresponding QA
boundary and does not directly alter preceding interaction events.
Cross-variant differences in pre-QA interaction metrics are therefore
descriptive whole-run differences, rather than isolated causal
effects of retrieval or ledger maintenance.

\subsection{Training Objectives}

Training uses session-disjoint event and QA records. The controller is trained to predict operation, addressee, completeness, and confidence:
\begin{equation}
\begin{aligned}
\mathcal{L}_{\mathrm{ctrl}}
&=
\mathcal{L}_{\mathrm{op}}^{\mathrm{wCE}}
+0.3\mathcal{L}_{\mathrm{addr}}^{\mathrm{CE}}
+0.3\mathcal{L}_{\mathrm{complete}}^{\mathrm{CE}}
+0.2\mathcal{L}_{\mathrm{conf}}^{\mathrm{BCE}} .
\end{aligned}
\label{eq:actmem_controller_loss}
\end{equation}
The weighted operation loss uses weights \((1.0,1.5,1.4,0.8)\) for \(\textsc{Write}\), \(\textsc{Update}\), \(\textsc{Ignore}\), and \(\textsc{Retain}\), respectively.

The retriever is trained with binary relevance supervision over active, superseded, forbidden, hard-negative, and distractor memory items:
\begin{equation}
\mathcal{L}_{\mathrm{retr}}
=
\mathcal{L}_{\mathrm{BCE}}(\mathbf{s},\mathbf{y})
+
\frac{0.5}{|\mathcal{P}||\mathcal{N}|}
\sum_{p\in\mathcal{P}}
\sum_{n\in\mathcal{N}}
\left[0.25-s_p+s_n\right]_{+},
\label{eq:actmem_retriever_loss}
\end{equation}
where \(\mathbf{s}\) denotes candidate scores, \(\mathbf{y}\) denotes binary relevance labels, and \(\mathcal{P}\) and \(\mathcal{N}\) denote positive and negative candidate sets.

The total objective is
\begin{equation}
\mathcal{L}_{\mathrm{ACT\text{-}Mem}}
=
\mathcal{L}_{\mathrm{ctrl}}
+
\mathcal{L}_{\mathrm{retr}}.
\label{eq:actmem_total_loss}
\end{equation}

\subsection{Optimization and Data Splits}

The training data are divided by session into train, validation, calibration, and held-out partitions containing 660, 175, 40, and 219 sessions, respectively. The controller and retriever are jointly optimized in the first stage using AdamW with learning rate \(3\times10^{-4}\), batch size 32, ten training epochs, weight decay \(0.01\), gradient-norm clipping at \(1.0\), and random seed \(20260907\).

In the second stage, the controller is fixed and the retriever is adapted with learning rate \(1\times10^{-4}\), batch size 32, three adaptation epochs, and random seed \(20260908\). Checkpoints are selected using only the training and validation partitions. The 40-session calibration split is used for threshold selection without gradient updates. Held-out sessions are reserved for the final comparison and are not used for parameter, checkpoint, or threshold selection.

Training is performed on eight NVIDIA A100 GPUs. The \texttt{MiniCPM}-o~4.5 backbone and the speech transcription and timing components remain frozen during both optimization stages.

\subsection{Calibration and Same-Session Inference}

Calibration evaluates candidate admission thresholds on the calibration sessions and selects the threshold that maximizes memory-eligibility F1. The selected threshold is then fixed for held-out inference. During a test session, the controller processes events in chronological order and updates the ledger after each event. The retriever is invoked only when a memory question reaches the stable-silence boundary.

The inference sequence for a memory question is
\begin{equation}
\begin{aligned}
\text{stable assistant silence}
&\rightarrow \text{question audio}
\rightarrow \text{typed retrieval} \\
&\rightarrow \text{evidence-card construction}
\rightarrow \text{native text prefill}
\rightarrow \text{spoken answer}.
\end{aligned}
\label{eq:actmem_qa_inference}
\end{equation}
The question audio and the evidence card are provided within the same persistent backbone session. Each question receives an independently constructed card, and the resulting answer is attributed to its own QA window.

\subsection{Runtime Artifacts and Audit Records}

For every evaluated session, \textsf{ACT-Mem} records the controller trace, ledger snapshot, retrieved memory identifiers, superseded identifiers, query type, retrieval scores, gate decision, evidence card, answer transcript, response/no-response status, and runtime latency. Each event trace contains the event identifier, event index, predicted operation, addressee, completeness, confidence, admission score, created slots, updated slots, superseded slots, and uncertain or aborted slots.

These records provide an auditable correspondence between an input event, its predicted memory operation, the resulting versioned state, the evidence shown at QA time, and the final spoken answer. They also preserve the effective session and question denominators used for the held-out evaluation.

\section{Training, Calibration, and Computational Cost}
\label{app:training_details}

\subsection{Training Setup}

\textsf{ACT-Mem} is trained with session-disjoint data partitions containing \(660\) training sessions, \(175\) validation sessions, \(40\) calibration sessions, and \(219\) held-out sessions. The \texttt{MiniCPM}-o~4.5 full-duplex backbone and the speech transcription and timing components remain frozen. Only the event controller and typed retriever are optimized, resulting in \(8.92\)M trainable parameters: \(4.59\)M in the controller and \(4.33\)M in the retriever.

Training consists of two stages. The first stage jointly optimizes the event controller and the typed retriever. The second stage initializes from the joint checkpoint, freezes the event controller, and further adapts the typed retriever using session-level memory evidence. Optimization uses AdamW with weight decay \(0.01\), batch size \(32\), and gradient-norm clipping at \(1.0\). The joint stage uses learning rate \(3\times10^{-4}\), ten epochs, and random seed \(20260907\); the adaptation stage uses learning rate \(1\times10^{-4}\), three epochs, and random seed \(20260908\). Training is performed on eight NVIDIA A100 GPUs.

\begin{table}[t]
\centering
\scriptsize
\caption{Training configuration.}
\label{tab:actmem_training_configuration}
\begin{tabular}{p{0.48\linewidth}l}
\toprule
\textbf{Configuration} & \textbf{Value} \\
\midrule
Session split & \(660/175/40/219\) train/validation/calibration/held-out \\
Trainable parameters & \(8.92\)M \\
Controller parameters & \(4.59\)M \\
Retriever parameters & \(4.33\)M \\
Optimization stages & Joint training followed by retriever adaptation \\
Optimizer & AdamW \\
Batch size & \(32\) \\
Weight decay & \(0.01\) \\
Gradient clipping & \(1.0\) \\
Joint-stage learning rate and epochs & \(3\times10^{-4}\), \(10\) epochs \\
Adaptation-stage learning rate and epochs & \(1\times10^{-4}\), \(3\) epochs \\
Training hardware & \(8\) NVIDIA A100 GPUs \\
\bottomrule
\end{tabular}
\end{table}

All supervision is organized by session, and the held-out sessions are excluded from training, checkpoint selection, and calibration. The final checkpoint is loaded jointly with the frozen speech backbone during same-session evaluation.

\subsection{Calibration Protocol}

Calibration is performed on the dedicated \(40\)-session calibration split without gradient updates. We search over candidate admission thresholds and select the value that maximizes event-memory eligibility F1. The resulting write threshold is
\[
\tau_{\mathrm{write}}=0.35.
\]
After calibration, the threshold is fixed for held-out inference. The calibration record stores the candidate thresholds, the selected value, the calibration split identifier, and the associated checkpoint identity.

\subsection{Computational Cost}

The computational cost reported for \textsf{ACT-Mem} is incremental relative to the frozen \texttt{MiniCPM}-o~4.5 backbone. The additional trainable module contains \(8.92\)M parameters, divided between the event controller and the typed retriever as shown in Table~\ref{tab:actmem_training_configuration}. During inference, the controller performs one chronological state update for each observed event, and the retriever constructs at most one evidence card for each memory question. The resulting card is inserted through the native question-conditioning path while the full-duplex backbone remains unchanged.

Runtime artifacts record the loaded checkpoint, effective calibration threshold, controller traces, retrieved memory identifiers, evidence cards, answer windows, and per-session failures. These records provide the basis for reporting the incremental memory-module cost separately from the fixed cost of full-duplex backbone inference.

\section{Statistical Analysis}
\label{app:statistics}

\subsection{Unit of Analysis and Aggregation}

The session is the primary statistical unit. The six memory questions within a session, as well as the interaction events belonging to that session, are first aggregated into session-level quantities. Overall results are then computed by arithmetic macro-averaging over sessions:
\begin{equation}
\overline{m}
=
\frac{1}{|\mathcal{I}_m|}
\sum_{i\in\mathcal{I}_m} m_i,
\label{eq:statistical_macro_average}
\end{equation}
where \(m_i\) is the value of metric \(m\) for session \(i\), and \(\mathcal{I}_m\) is the set of sessions with a valid value for that metric.

Questions from the same session are not treated as independent statistical replicates. Failure handling is metric-specific. Unanswered memory questions
receive no credit under fixed question denominators, and unrealized
challenges remain in the planned-challenge denominator for RCR.
Interaction metrics use their eligible-event denominators. A session
with no valid denominator for a metric is excluded from that
metric-specific mean, with the effective denominator reported;
undefined values are not replaced by zero.

\subsection{Paired Session Bootstrap}

The controlled comparison between \textsf{ACT-Mem} and the frozen \texttt{MiniCPM}-o~4.5 backbone uses the same \(219\) held-out sessions for both systems. We therefore use a paired percentile bootstrap at the session level. For each bootstrap replicate \(b\), \(219\) session identifiers are sampled with replacement. The outputs of both systems for each sampled identifier are kept paired, and the session-level metric is recomputed before aggregation.

For metric \(m\), the paired effect is
\begin{equation}
\Delta_m
=
\overline{m}_{\mathrm{ACT\text{-}Mem}}
-
\overline{m}_{\mathrm{MiniCPM}},
\label{eq:paired_effect}
\end{equation}
where the two means are computed on the same effective set of sessions for that metric. We generate \(20{,}000\) bootstrap replicates using random seed \(20260807\). The \(95\%\) percentile confidence interval is
\begin{equation}
\mathrm{CI}_{95}(m)
=
\left[
Q_{0.025}\!\left(\{\Delta_m^{(b)}\}_{b=1}^{20000}\right),
Q_{0.975}\!\left(\{\Delta_m^{(b)}\}_{b=1}^{20000}\right)
\right],
\label{eq:paired_bootstrap_ci}
\end{equation}
where \(Q_{\alpha}\) denotes the empirical \(\alpha\)-quantile of the bootstrap effect distribution.

The primary inferential intervals are reported for latest-value accuracy and final-state fact coverage. When intervals are available for additional memory or interaction metrics, they are computed using the same paired session-level procedure. Absolute changes in percentage points are reported as the primary effect size; relative percentage changes are descriptive summaries.

Pairing is by session identifier and does not imply identical
realized challenge delivery across systems. The intervals quantify
session-sampling uncertainty conditional on the recorded runs
and the specified scorer; they do not account for repeated-run
generation variability, ASR errors, or scoring-rule validity.

\subsection{Baseline and Subgroup Reporting}

The results for the five full-duplex baseline models are primarily descriptive. Each model is evaluated on the same session manifests and is aggregated with the same session-level procedure. No scalar ranking is derived by combining interaction and memory metrics. In particular, challenge realization is interpreted as exposure, while conditional full-duplex scores are interpreted together with challenge realization, trigger-timeout rate, and behavioral observability.

Language-, scenario-, and question-type analyses are reported as exploratory stratified summaries. The same definitions, denominators, and macro-aggregation rule are used within each stratum. Each subgroup result is accompanied by its session count and effective metric denominator. These stratified analyses are not treated as independent confirmatory tests, and no additional significance claims are made without a prespecified multiple-comparison procedure.

\subsection{Denominators and Reproducibility}

For every reported metric, we retain the attempted-session count and the effective denominator used for aggregation. Runtime metrics additionally retain protocol-valid and behaviorally observable session counts, while memory metrics retain the total number of questions and the number of questions of each relevant type. Conditional quantities, including conditional full-duplex behavior and answer accuracy given a response, are reported together with their coverage variables.

All bootstrap computations use the fixed session identifiers, the fixed held-out split, and the fixed random seed specified above. The released statistical records contain session-level metric values, effective denominators, paired differences, bootstrap settings, and confidence intervals, allowing the reported aggregate results to be reproduced without treating individual questions as independent observations.

\section{Complete Benchmark Results}
\label{app:complete_results}

\subsection{Evaluation Scope}

This section reports the complete aggregate results for \texttt{MiniCPM}-o~4.5, \texttt{Freeze-Omni}, \texttt{Moshi}, \texttt{Lychee-FD}, and \texttt{PersonaPlex}-7B-v1 on all \(1{,}094\) sessions. Each model is evaluated on the same immutable session manifests, comprising \(6{,}564\) same-session memory questions and checks, including \(2{,}188\) latest-value questions. All ratios are calculated within each session and then macro-averaged over sessions. The tables below include all aggregate runtime and memory metrics defined in Appendix~\ref{app:metrics}.

\subsection{Runtime Validity and Challenge Exposure}

Table~\ref{tab:complete_runtime_exposure} reports execution validity, behavioral observability, adaptive-challenge exposure, conditional full-duplex behavior, response coverage, and assistant speech duration. Challenge realization measures the fraction of planned challenges that became behaviorally realized. Conditional cFD evaluates behavior on scorable challenges, whereas Adaptive FD is the macro-average of the corresponding session-level adaptive score. Assistant speech duration is descriptive and is not treated as a performance measure.

\begin{table}[t]
\centering
\caption{Runtime validity, challenge exposure, and response coverage on all \(1{,}094\) sessions. Values are percentages except assistant speech duration.}
\label{tab:complete_runtime_exposure}
\scriptsize
\setlength{\tabcolsep}{2.2pt}
\renewcommand{\arraystretch}{1.18}
\begin{tabularx}{\textwidth}{@{}l*{8}{>{\centering\arraybackslash}X}@{}}
\toprule
\textbf{Model}
& \shortstack{\textbf{Protocol}\\\textbf{valid}\(\uparrow\)}
& \shortstack{\textbf{Behaviorally}\\\textbf{observable}\(\uparrow\)}
& \shortstack{\textbf{Challenge}\\\textbf{realization}\(\uparrow\)}
& \shortstack{\textbf{Conditional}\\\textbf{cFD}\(\uparrow\)}
& \shortstack{\textbf{Adaptive}\\\textbf{FD}\(\uparrow\)}
& \shortstack{\textbf{Trigger}\\\textbf{timeout}\(\downarrow\)}
& \shortstack{\textbf{Response}\\\textbf{coverage}\(\uparrow\)}
& \shortstack{\textbf{Speech}\\\textbf{duration (s)}} \\
\midrule
\texttt{MiniCPM}-o~4.5 & 100.00 & 100.00 & 73.51 & 40.23 & 35.28 & 6.04 & 28.38 & 78.35 \\
\texttt{Freeze-Omni} & 100.00 & 100.00 & 86.17 & 28.40 & 30.41 & 0.53 & 76.03 & 203.16 \\
\texttt{Moshi} & 100.00 & 100.00 & 44.63 & 37.36 & 13.08 & 27.67 & 12.61 & 29.23 \\
\texttt{Lychee-FD} & 99.82 & 100.00 & 75.64 & 47.32 & 48.19 & 1.09 & 35.30 & 99.19 \\
\texttt{PersonaPlex}-7B-v1 & 100.00 & 100.00 & 93.54 & 22.74 & 35.41 & 1.73 & 32.43 & 74.20 \\
\bottomrule
\end{tabularx}

\vspace{1mm}
\parbox{\linewidth}{\raggedright\scriptsize \textit{Notes.} Conditional cFD is conditioned on behaviorally scorable challenges and must be interpreted together with challenge realization and trigger timeout. Adaptive FD is aggregated from session-level scores and therefore need not equal the product of the two separately aggregated columns.}
\end{table}

\subsection{Turn-Boundary and Interruption Metrics}

Table~\ref{tab:complete_runtime_timing} reports ordinary-turn boundary behavior, pause intrusion, and interruption stopping. Premature onset measures assistant speech beginning before the valid user-end boundary. Carry-in indicates that assistant speech from the preceding response remains active when the next user input begins. Pre-end overlap records any assistant speech overlapping the user before the user-end boundary.

\begin{table}[t]
\centering
\caption{Turn-boundary and interruption-handling results. Rates are percentages and stopping latencies are measured in seconds.}
\label{tab:complete_runtime_timing}
\scriptsize
\setlength{\tabcolsep}{2.8pt}
\renewcommand{\arraystretch}{1.18}
\begin{tabularx}{\textwidth}{@{}l*{7}{>{\centering\arraybackslash}X}@{}}
\toprule
\textbf{Model}
& \shortstack{\textbf{Premature}\\\textbf{onset}\(\downarrow\)}
& \shortstack{\textbf{Carry-}\\\textbf{in}\(\downarrow\)}
& \shortstack{\textbf{Pre-end}\\\textbf{overlap}\(\downarrow\)}
& \shortstack{\textbf{Pause}\\\textbf{intrusion}\(\downarrow\)}
& \shortstack{\textbf{Stop}\\\textbf{P50}\(\downarrow\)}
& \shortstack{\textbf{Stop}\\\textbf{P90}\(\downarrow\)}
& \shortstack{\textbf{Stop@}\\\textbf{0.8\,s}\(\uparrow\)} \\
\midrule
\texttt{MiniCPM}-o~4.5 & 18.70 & 25.86 & 44.38 & 59.47 & 1.032 & 1.446 & 55.69 \\
\texttt{Freeze-Omni} & 49.76 & 42.44 & 91.82 & 73.64 & 1.239 & 1.802 & 45.75 \\
\texttt{Moshi} & 66.59 & 5.49 & 72.25 & 45.11 & 0.777 & 0.864 & 60.79 \\
\texttt{Lychee-FD} & 24.68 & 32.96 & 57.14 & 53.89 & 0.312 & 0.440 & 90.12 \\
\texttt{PersonaPlex}-7B-v1 & 46.64 & 26.23 & 72.73 & 90.62 & 1.689 & 2.104 & 16.40 \\
\bottomrule
\end{tabularx}
\end{table}

\subsection{Same-Session Memory Metrics}

Tables~\ref{tab:complete_memory_updates} and~\ref{tab:complete_memory_selectivity} report all aggregate memory metrics. Latest-value accuracy requires the updated value to be present and the superseded value to be absent. Answer accuracy given response evaluates the same criterion only among latest-value questions for which an answer was produced.

\begin{table}[t]
\centering
\caption{Response availability and factual-update results. All values are percentages.}
\label{tab:complete_memory_updates}
\scriptsize
\setlength{\tabcolsep}{3.2pt}
\renewcommand{\arraystretch}{1.18}
\begin{tabularx}{\textwidth}{@{}l*{5}{>{\centering\arraybackslash}X}@{}}
\toprule
\textbf{Model}
& \shortstack{\textbf{Response}\\\textbf{rate}\(\uparrow\)}
& \shortstack{\textbf{No-response}\\\textbf{rate}\(\downarrow\)}
& \shortstack{\textbf{Latest-value}\\\textbf{accuracy}\(\uparrow\)}
& \shortstack{\textbf{Superseded-value}\\\textbf{error}\(\downarrow\)}
& \shortstack{\textbf{Accuracy given}\\\textbf{response}\(\uparrow\)} \\
\midrule
\texttt{MiniCPM}-o~4.5 & 98.90 & 1.10 & 4.30 & 0.59 & 4.32 \\
\texttt{Freeze-Omni} & 99.94 & 0.06 & 11.61 & 3.38 & 11.61 \\
\texttt{Moshi} & 54.31 & 45.69 & 2.19 & 0.32 & 3.60 \\
\texttt{Lychee-FD} & 92.41 & 7.59 & 25.50 & 1.65 & 30.80 \\
\texttt{PersonaPlex}-7B-v1 & 96.40 & 3.60 & 5.85 & 2.29 & 5.93 \\
\bottomrule
\end{tabularx}
\end{table}

\begin{table}[t]
\centering
\caption{Distractor handling, final-state coverage, and answer-overlap results. All values are percentages.}
\label{tab:complete_memory_selectivity}
\scriptsize
\setlength{\tabcolsep}{4.0pt}
\renewcommand{\arraystretch}{1.18}
\begin{tabularx}{\textwidth}{@{}l*{4}{>{\centering\arraybackslash}X}@{}}
\toprule
\textbf{Model}
& \shortstack{\textbf{Distractor}\\\textbf{adoption}\(\downarrow\)}
& \shortstack{\textbf{Distractor}\\\textbf{rejection}\(\uparrow\)}
& \shortstack{\textbf{Summary fact}\\\textbf{coverage}\(\uparrow\)}
& \shortstack{\textbf{Legacy token}\\\(\boldsymbol{F_1}\)\(\uparrow\)} \\
\midrule
\texttt{MiniCPM}-o~4.5 & 14.17 & 85.19 & 2.01 & 29.45 \\
\texttt{Freeze-Omni} & 65.90 & 34.10 & 0.00 & 15.42 \\
\texttt{Moshi} & 44.15 & 8.32 & 0.04 & 7.64 \\
\texttt{Lychee-FD} & 48.17 & 51.55 & 1.06 & 21.36 \\
\texttt{PersonaPlex}-7B-v1 & 79.71 & 17.46 & 0.24 & 7.70 \\
\bottomrule
\end{tabularx}

\vspace{1mm}
\parbox{\linewidth}{\raggedright\scriptsize \textit{Notes.} Distractor adoption and rejection are reported together because silence may reduce adoption without demonstrating correct rejection. Legacy token \(F_1\) is retained for compatibility and is not used as the sole measure of memory fidelity.}
\end{table}

\subsection{Effective Denominators}

Table~\ref{tab:complete_result_denominators} reports the principal aggregate denominators. All models are evaluated on \(1{,}094\) attempted sessions. \texttt{Lychee-FD} contains \(1{,}092\) protocol-valid sessions; its remaining behaviorally observable sessions are retained in the attempted-session accounting. Metric-specific denominators for challenge, timing, pause, and interruption statistics follow the definitions in Appendix~\ref{app:metrics}.

\begin{table}[t]
\centering
\caption{Principal denominators for the complete benchmark evaluation.}
\label{tab:complete_result_denominators}
\small
\setlength{\tabcolsep}{5.0pt}
\renewcommand{\arraystretch}{1.18}
\begin{tabular}{lrrrrr}
\toprule
\textbf{Model}
& \shortstack{\textbf{Attempted}\\\textbf{sessions}}
& \shortstack{\textbf{Protocol-valid}\\\textbf{sessions}}
& \shortstack{\textbf{Observable}\\\textbf{sessions}}
& \shortstack{\textbf{Memory}\\\textbf{questions}}
& \shortstack{\textbf{Latest-value}\\\textbf{questions}} \\
\midrule
\texttt{MiniCPM}-o~4.5 & 1,094 & 1,094 & 1,094 & 6,564 & 2,188 \\
Freeze-Omni & 1,094 & 1,094 & 1,094 & 6,564 & 2,188 \\
\texttt{Moshi} & 1,094 & 1,094 & 1,094 & 6,564 & 2,188 \\
\texttt{Lychee-FD} & 1,094 & 1,092 & 1,094 & 6,564 & 2,188 \\
\texttt{PersonaPlex}-7B-v1 & 1,094 & 1,094 & 1,094 & 6,564 & 2,188 \\
\bottomrule
\end{tabular}
\end{table}

\section{Qualitative Examples and Failure Analysis}
\label{app:qualitative}

\subsection{Example Selection and Presentation}

We present representative traces that expose failure modes not fully visible from aggregate scores. The examples were selected from completed, protocol-valid evaluation sessions and were not used for training, calibration, or checkpoint selection.

Each trace combines the scheduled challenge, the expected state transition, runtime interaction measurements, and the model's spoken answer. Spoken outputs are reported from the frozen-ASR transcripts, with only whitespace and terminal punctuation normalized for readability. The examples are diagnostic rather than estimates of failure frequency.

\subsection{Interaction and Memory Can Fail Independently}

Table~\ref{tab:qualitative_scenario_examples} summarizes examples covering the four full-duplex scenarios. The traces demonstrate that correct interaction timing does not guarantee coherent memory and, conversely, that recovering an isolated fact does not imply correct full-duplex behavior.

\begin{table*}[t]
\centering
\begin{threeparttable}
\caption{Representative qualitative examples from protocol-valid sessions. RCR and cFD are reported as proportions for the selected session.}
\label{tab:qualitative_scenario_examples}
\scriptsize
\setlength{\tabcolsep}{3.2pt}
\renewcommand{\arraystretch}{1.16}

\begin{tabularx}{\textwidth}{
@{}
>{\raggedright\arraybackslash}p{0.105\textwidth}
>{\raggedright\arraybackslash}p{0.205\textwidth}
>{\raggedright\arraybackslash}X
>{\raggedright\arraybackslash}p{0.245\textwidth}
@{}}
\toprule
\textbf{Scenario} & \textbf{Expected behavior} & \textbf{Observed evidence} & \textbf{Diagnosis} \\
\midrule
Interruption & Stop the current response, replace \emph{Saturday evening} with \emph{Saturday morning}, and replace \emph{twenty-five minutes} with \emph{ten minutes}. & Both interruptions were realized, with median stopping latency $0.368$\,s and Stop@0.8 of $1.00$. The model recovered \emph{Saturday morning}, but did not recover \emph{ten minutes}. When asked about the side information, it answered that the distractor should be saved. & Fast interruption stopping was accompanied by only partial factual revision and incorrect admission. This separates acoustic interruption recovery from transactional memory updating. \\

\addlinespace[1mm]

Pause handling & Remain silent during within-turn pauses, then commit the completed correction: \emph{Thursday morning} and \emph{fifteen minutes}. & For the same session, both examined models intruded into every evaluated pause. One model rejected the distractor but recovered neither scored latest value. The other recovered \emph{Thursday morning}, but answered ``I will save all the details into the final memory'' for the distractor question and later included a \emph{travel toothbrush} in its summary. & The first behavior is a missed update; the second is an admission failure followed by summary contamination. Both occur together with incorrect pause-boundary control. \\

\addlinespace[1mm]

Backchannel & Continue the current response without treating the backchannel as a new turn or a persistent fact. & All three challenges were realized, but session cFD was $0.33$. The model nevertheless recalled the stable requirement to avoid hazelnut flavoring and retain whole beans, and correctly stated that the kitchen-clutter detail should not be saved. Its final summary remained incomplete. & Correct retention and distractor rejection can coexist with poor continuation behavior. Backchannel handling and final-state consolidation therefore require separate attribution. \\

\addlinespace[1mm]

Speech to others & Do not answer the non-addressed speech and do not admit its content into memory. & The session obtained RCR $0.75$ and cFD $0.25$. The model explicitly rejected the side information, but failed both latest-value questions and produced an incomplete final summary. & Rejecting a queried distractor does not establish that the system maintained the intended active state. Addressee control, supersession, and later answer generation remain distinct stages. \\
\bottomrule
\end{tabularx}

\begin{tablenotes}[flushleft]
\scriptsize
\item[] Quoted outputs are frozen-ASR transcriptions. The interpretations use the same fixed-denominator decisions as the quantitative evaluation; no qualitative relabeling is applied.
\end{tablenotes}
\end{threeparttable}
\end{table*}

\paragraph{Interruption is not equivalent to revision.}
The interruption example shows the clearest interaction--memory dissociation. The model stopped within the required interval for both realized challenges, yielding perfect stopping success in the selected session. However, it recovered only one of the two corrected values and admitted information explicitly marked as irrelevant. The interruption mechanism therefore reacted to the acoustic event without constructing the complete post-correction state.

\paragraph{Pause errors can precede otherwise plausible answers.}
In the pause-handling example, both models spoke during a region that should have been treated as an unfinished user turn. Their later memory behaviors diverged: one was selective but unable to state the revised values, whereas the other recovered a revised value but admitted a personal item. This case illustrates why a plausible answer to one question cannot substitute for joint measurement of interaction boundaries and persistent-state fidelity.

\paragraph{Explicit rejection is stronger evidence than silence.}
A model receives distractor-rejection credit only when its spoken answer explicitly rejects the irrelevant information. No response remains an unanswered case. This distinction prevents systems with low response coverage from appearing selective merely because they remain silent.

\subsection{Auditable Transaction Trace}

Table~\ref{tab:actmem_qualitative_trace} presents an \textsf{ACT-Mem} trace from a held-out English pause-handling session. The user first supplied provisional budgeting values, later corrected the review time to \emph{this Saturday morning} and the electricity cap to \emph{\$150}, and explicitly marked an unrelated detail as unsuitable for storage.

\begin{table*}[t]
\centering
\begin{threeparttable}
\caption{An \textsf{ACT-Mem} transaction trace illustrating both successful control decisions and downstream state-construction failures.}
\label{tab:actmem_qualitative_trace}
\scriptsize
\setlength{\tabcolsep}{3.2pt}
\renewcommand{\arraystretch}{1.15}

\begin{tabularx}{\textwidth}{
@{}
>{\centering\arraybackslash}p{0.075\textwidth}
>{\raggedright\arraybackslash}p{0.205\textwidth}
>{\raggedright\arraybackslash}p{0.225\textwidth}
>{\raggedright\arraybackslash}X
@{}}
\toprule
\textbf{Stage} & \textbf{Gold transition} & \textbf{Predicted transaction} & \textbf{Consequence} \\
\midrule
Initial request & \textsc{Write} provisional review time and budget constraints. & The controller predicted \textsc{Write} and committed extracted entity--value slots. & The initial values became active and were available to subsequent retrieval. \\

Correction & \textsc{Update} the review time to \emph{this Saturday morning} and the electricity cap to \emph{\$150}; supersede both provisional values. & The controller predicted \textsc{Update}, but the extracted correction was represented as one compound slot with an entity key different from the provisional slots. & A new item was created, while the original items remained active and no supersession link was formed. The failure therefore occurred after the operation decision, during slot normalization and ledger resolution. \\

Distractor-bearing event & Commit the relevant condition and \textsc{Ignore} the explicitly excluded side detail. & The relevant condition was written, but the exclusion clause ``just a side detail and should not be saved'' was also extracted as an active fact. & The ledger became contaminated even though the user's utterance explicitly expressed non-persistence. This is an incorrect-admission failure rather than a retrieval-only failure. \\

Latest-value QA & Retrieve the active corrected electricity cap and answer \emph{\$150}. & No spoken answer was produced. & The case remains incorrect under the fixed denominator. The trace localizes the outcome to unresolved state construction followed by answer realization. \\

Distractor QA & Produce an explicit rejection without exposing the side detail as evidence. & No spoken answer was produced. & Neither adoption nor successful rejection is credited; silence is kept separate from selective exclusion. \\
\bottomrule
\end{tabularx}

\begin{tablenotes}[flushleft]
\scriptsize
\item[] Controller decisions and slot contents are derived from frozen-ASR event transcripts. The trace reports the stored state before answer scoring and therefore permits failures to be localized without inferring hidden state from the spoken answer alone.
\end{tablenotes}
\end{threeparttable}
\end{table*}

This trace illustrates the value of retaining transaction provenance. The high-level \textsc{Update} decision was correct, but the correction did not supersede the provisional entries because the extracted entity representation was inconsistent. An answer-only evaluation would record a failed latest-value question without distinguishing a missed correction from failed entity resolution, failed supersession, retrieval error, or response omission.

The same trace also reveals a limitation of clause-level admission. A negated exclusion statement may contain a fluent entity--value pattern, causing its literal wording to be stored even though its pragmatic function is to prohibit storage. More robust semantic scope modeling is therefore required for utterances that combine relevant facts, corrections, and explicit non-persistence instructions.

\subsection{Failure Taxonomy}

We assign each qualitative error to the earliest observable stage that can explain it. Table~\ref{tab:qualitative_failure_taxonomy} defines the resulting taxonomy. Later symptoms are retained for analysis but are not treated as evidence that all earlier stages were correct.

\begin{table}[t]
\centering
\caption{Failure taxonomy used for qualitative attribution.}
\label{tab:qualitative_failure_taxonomy}
\scriptsize
\setlength{\tabcolsep}{3.5pt}
\renewcommand{\arraystretch}{1.12}

\begin{tabularx}{\linewidth}{
@{}
>{\raggedright\arraybackslash}p{0.28\linewidth}
>{\raggedright\arraybackslash}X
@{}}
\toprule
\textbf{Failure} & \textbf{Observable criterion} \\
\midrule
Trigger timeout & A speech-dependent challenge is planned but cannot be injected because the required assistant-speech condition is not reached. \\
Boundary violation & The assistant begins before a completed user turn, speaks during a protected pause, or intrudes into a non-response window. \\
Slow interruption stop & Assistant speech continues beyond the stopping threshold after an interruption is delivered. \\
Missed write & An admissible new fact produces no committed memory item. \\
Incorrect admission & Non-addressed, incomplete, backchannel, or explicitly excluded content is committed as persistent memory. \\
Failed supersession & A valid correction is detected, but the prior active value is not linked and marked as superseded. \\
Stale retrieval & A superseded value is exposed to the answer generator despite the presence of a valid active replacement. \\
Distractor leakage & An excluded item enters the evidence card or appears affirmatively in the spoken answer or final summary. \\
Retrieval miss & The required active item exists in the ledger but is absent from the question-conditioned evidence. \\
Answer realization error & Appropriate evidence is available, but the spoken response omits, distorts, or contradicts it, or no answer is produced. \\
Transcription error & The frozen-ASR transcript changes a value or boundary in a way that affects controller input or answer scoring. \\
\bottomrule
\end{tabularx}
\end{table}

The examples identify three recurring patterns. First, interaction success is local: rapid stopping or correct backchannel continuation does not guarantee admission or revision of the corresponding facts. Second, memory selectivity is compositional: a system may reject one queried distractor while retaining stale values or producing an incomplete summary. Third, an explicit transaction trace makes compound errors separable. In particular, correct operation classification does not ensure correct slot extraction, entity resolution, supersession, retrieval, or spoken realization. These distinctions explain why the benchmark reports interaction boundaries and memory outcomes separately and why no single scalar score is sufficient to characterize full-duplex persistent memory.

\section{Reproducibility and Artifact Manifest}
\label{app:artifacts}

This section specifies the artifact organization required to reproduce the reported experiments. All released records use stable session identifiers and preserve the mapping from each input session to its events, challenges, questions, model outputs, and scores. The artifact structure is summarized in Table~\ref{tab:artifact_manifest}.

\subsection{Artifact Organization}

\begin{table}[t]
\centering
\caption{Artifact manifest for reproducing the benchmark and experiments.}
\label{tab:artifact_manifest}
\scriptsize
\setlength{\tabcolsep}{3pt}
\renewcommand{\arraystretch}{1.14}

\begin{tabularx}{\linewidth}{
@{}
>{\raggedright\arraybackslash}p{0.22\linewidth}
>{\raggedright\arraybackslash}p{0.27\linewidth}
>{\raggedright\arraybackslash}X
@{}}
\toprule
\textbf{Artifact group} & \textbf{Representative records} & \textbf{Reproducibility role} \\
\midrule
Session inputs & \texttt{source.json}, session audio, event audio & Defines language, scenario, event order, speaker roles, timing, and the rendered input waveform. \\
Interaction manifests & \texttt{interactive\_manifest.json}, challenge and timing manifests & Specifies challenge conditions, observation windows, trigger rules, and the fixed execution order. \\
Memory supervision & \texttt{memory\_qa.json}, reference-state records, update links & Provides question types, accepted and forbidden values, expected operations, supporting evidence, and final-state supervision. \\
Reference behavior & \texttt{reference\_behavior.json} & Stores the intended interaction response for each challenge and the corresponding state-transition constraints. \\
Runtime execution & \texttt{runtime\_events.jsonl}, VAD and timing records & Records challenge realization, response boundaries, interruption stopping, overlap, pauses, and execution failures. \\
Generated outputs & Assistant waveform, raw predictions, QA windows, frozen-ASR transcripts & Preserves the model output and the temporal windows used for answer attribution and transcription. \\
Question-level scores & Per-question score records and session summaries & Allows reconstruction of response coverage, latest-value accuracy, stale assertions, distractor outcomes, and final-state coverage. \\
\textsf{ACT-Mem} traces & Controller--ledger traces, retrieval records, evidence cards & Exposes event operations, committed items, supersession links, active values, and question-conditioned evidence. \\
Configuration and scripts & Inference, scoring, aggregation, and calibration configurations & Reproduces execution settings and regenerates the reported tables from the recorded outputs. \\
\bottomrule
\end{tabularx}
\end{table}

The records are linked by session identifier and question identifier. Consequently, a question-level score can be traced back to its input audio, the corresponding event prefix, the realized challenge schedule, the generated answer, and the evidence used for scoring. Failed, unanswered, and timed-out executions remain in the released records and are not removed before aggregation.

\subsection{Data Splits and Provenance}

The release-level development and test partitions contain 237 and 857 sessions, respectively. The method-development partitions contain 660 training sessions, 175 validation sessions, 40 calibration sessions, and 219 held-out sessions. Splits are defined at the session level, so that events, questions, and audio derived from one session cannot appear in more than one partition.

The training and validation records are used for parameter fitting and checkpoint selection. The calibration records are used only for selecting the admission threshold. The held-out records are reserved for the final comparison and are not used for training, calibration, or model selection. Each reported held-out result can therefore be reconstructed from the corresponding fixed session manifest and its stored output records.

\subsection{Execution and Score Reconstruction}

The complete reproduction path is

\begin{equation}
\begin{split}
\text{session manifest}
&\rightarrow \text{streaming execution}
\rightarrow \text{runtime event log}\\
&\rightarrow \text{generated audio}
\rightarrow \text{frozen-ASR transcript}\\
&\rightarrow \text{question-level scoring}
\rightarrow \text{session aggregation}
\rightarrow \text{reported table}.
\end{split}
\label{eq:artifact_reproduction_path}
\end{equation}

The runtime log determines whether an adaptive challenge was realized and provides the timestamps required for interaction metrics. The question records determine the fixed denominators for memory evaluation. Session ratios are computed before macro-averaging across sessions, and unanswered or invalid responses remain in the corresponding denominators. The final aggregation files are generated from these session-level records rather than manually transcribed into the tables.

For the controlled comparison, the held-out session identifiers are retained in paired form for the backbone and memory-augmented runs. This allows the reported confidence intervals to be regenerated with paired session resampling while preserving the correspondence between the two conditions.

\subsection{Integrity and Reproduction Checks}

Before scoring, the reproduction script verifies that every session manifest has a matching input waveform, event list, challenge record, and question set. It also checks that event and question references are unique and resolvable, that all event timestamps fall within the declared session duration, and that no session identifier occurs in multiple partitions. After inference, the script verifies that every output can be mapped to a fixed input session and that the score denominators agree with the released question manifests.

These checks ensure that the reported results are reproducible from session-level inputs and recorded outputs, while preserving the traceability required to inspect individual interaction and memory failures.

\end{document}

%% file: math_commands.tex
\usepackage{amsmath,amsfonts,bm}

\def\eqref#1{equation~\ref{#1}}

\def\1{\bm{1}}

\DeclareMathAlphabet{\mathsfit}{\encodingdefault}{\sfdefault}{m}{sl}
\SetMathAlphabet{\mathsfit}{bold}{\encodingdefault}{\sfdefault}{bx}{n}



%% file: iclr2027_conference.bib
@misc{packer2023memgpt,
  title         = {{MemGPT}: Towards {LLM}s as Operating Systems},
  author        = {Packer, Charles and Fang, Vivian and Patil, Shishir G. and Lin, Kevin and Wooders, Sarah and Gonzalez, Joseph E.},
  year          = {2023},
  eprint        = {2310.08560},
  archivePrefix = {arXiv},
  primaryClass  = {cs.AI},
  url           = {https://arxiv.org/abs/2310.08560}
}

@inproceedings{xu2025amem,
  title     = {A-MEM: Agentic Memory for {LLM} Agents},
  author    = {Xu, Wujiang and Liang, Zujie and Mei, Kai and Gao, Hang and Tan, Juntao and Zhang, Yongfeng},
  booktitle = {Advances in Neural Information Processing Systems},
  year      = {2025},
  url       = {https://arxiv.org/abs/2502.12110}
}

@misc{yao2025egomem,
  title         = {EgoMem: Lifelong Memory Agent for Full-Duplex Omnimodal Models},
  author        = {Yao, Yiqun and others},
  year          = {2025},
  eprint        = {2509.11914},
  archivePrefix = {arXiv},
  primaryClass  = {cs.AI},
  url           = {https://arxiv.org/abs/2509.11914}
}

@misc{xie2026voicemem,
  title         = {VoiceMem: Streaming Memory for Long-Term Voice Interaction},
  author        = {Xie, Zhifei and others},
  year          = {2026},
  eprint        = {2608.26005},
  archivePrefix = {arXiv},
  primaryClass  = {cs.CL},
  url           = {https://arxiv.org/abs/2608.26005}
}

@inproceedings{lin2025fullduplexbench,
  title     = {Full-Duplex-Bench: A Benchmark to Evaluate Full-Duplex Spoken Dialogue Models on Turn-Taking Capabilities},
  author    = {Lin, Guan-Ting and Lian, Jiachen and Li, Tingle and Wang, Qirui and Anumanchipalli, Gopala and Liu, Alexander H. and Lee, Hung-yi},
  booktitle = {Proceedings of the IEEE Automatic Speech Recognition and Understanding Workshop},
  year      = {2025},
  url       = {https://arxiv.org/abs/2503.04721}
}

@inproceedings{lin2026fullduplexbenchv15,
  title     = {Full-Duplex-Bench v1.5: Evaluating Overlap Handling for Full-Duplex Spoken Dialogue Models},
  author    = {Lin, Guan-Ting and Kuan, Shih-Yun Shan and Wang, Qirui and Lian, Jiachen and Li, Tingle and Lee, Hung-yi},
  booktitle = {Proceedings of the IEEE International Conference on Acoustics, Speech and Signal Processing},
  year      = {2026},
  url       = {https://arxiv.org/abs/2507.23159}
}

@misc{wang2026humdial,
  title         = {{HumDial-FDBench}: A Human-Recorded Benchmark for Full-Duplex Spoken Dialogue},
  author        = {Wang, Chengyou and Xue, Hongfei and Li, Guojian and Zhao, Zhixian and Wang, Shuiyuan and Wang, Shuai and Xu, Xin and Bu, Hui and Xie, Lei},
  year          = {2026},
  eprint        = {2604.21406},
  archivePrefix = {arXiv},
  primaryClass  = {cs.CL},
  url           = {https://arxiv.org/abs/2604.21406}
}

@inproceedings{lin2026fullduplexbenchv2,
  title     = {Full-Duplex-Bench v2: An Adaptive Benchmark for Full-Duplex Spoken Dialogue Models},
  author    = {Lin, Guan-Ting and others},
  booktitle = {Proceedings of the 64th Annual Meeting of the Association for Computational Linguistics},
  year      = {2026},
  url       = {https://arxiv.org/abs/2510.07838}
}

@misc{lin2026fullduplexbenchv3,
  title         = {Full-Duplex-Bench v3: Evaluating Tool Use under Speech Disfluency},
  author        = {Lin, Guan-Ting and Chen, Chen and Chen, Zhehuai and Lee, Hung-yi},
  year          = {2026},
  eprint        = {2604.04847},
  archivePrefix = {arXiv},
  primaryClass  = {cs.CL},
  url           = {https://arxiv.org/abs/2604.04847}
}

@misc{modi2026echochain,
  title         = {EchoChain: Evaluating State Updates under Interruption in Full-Duplex Spoken Dialogue},
  author        = {Modi, Smit Nautambhai and Mahajan, Gandharv and Wetter, Marc and Welles, Randall},
  year          = {2026},
  eprint        = {2604.16456},
  archivePrefix = {arXiv},
  primaryClass  = {cs.CL},
  url           = {https://arxiv.org/abs/2604.16456}
}

@inproceedings{maharana2024locomo,
  title     = {Evaluating Very Long-Term Conversational Memory of {LLM} Agents},
  author    = {Maharana, Adyasha and Lee, Dong-Ho and Tulyakov, Sergey and Bansal, Mohit and Barbieri, Francesco and Fang, Yuwei},
  booktitle = {Proceedings of the 62nd Annual Meeting of the Association for Computational Linguistics},
  pages     = {13851--13870},
  year      = {2024},
  doi       = {10.18653/v1/2024.acl-long.747},
  url       = {https://aclanthology.org/2024.acl-long.747/}
}

@inproceedings{wu2025longmemeval,
  title     = {{LongMemEval}: Benchmarking Chat Assistants on Long-Term Interactive Memory},
  author    = {Wu, Di and others},
  booktitle = {International Conference on Learning Representations},
  year      = {2025},
  url       = {https://openreview.net/forum?id=pZiyCaVuti}
}

@inproceedings{karpukhin2020dpr,
  title     = {Dense Passage Retrieval for Open-Domain Question Answering},
  author    = {Karpukhin, Vladimir and Oguz, Barlas and Min, Sewon and Lewis, Patrick and Wu, Ledell and Edunov, Sergey and Chen, Danqi and Yih, Wen-tau},
  booktitle = {Proceedings of the 2020 Conference on Empirical Methods in Natural Language Processing},
  pages     = {6769--6781},
  year      = {2020},
  publisher = {Association for Computational Linguistics},
  doi       = {10.18653/v1/2020.emnlp-main.550},
  url       = {https://aclanthology.org/2020.emnlp-main.550/}
}

@inproceedings{lewis2020rag,
  title     = {Retrieval-Augmented Generation for Knowledge-Intensive {NLP} Tasks},
  author    = {Lewis, Patrick and Perez, Ethan and Piktus, Aleksandra and Petroni, Fabio and Karpukhin, Vladimir and Goyal, Naman and K{\"u}ttler, Heinrich and Lewis, Mike and Yih, Wen-tau and Rockt{\"a}schel, Tim and Riedel, Sebastian and Kiela, Douwe},
  booktitle = {Advances in Neural Information Processing Systems},
  volume    = {33},
  pages     = {9459--9474},
  year      = {2020},
  url       = {https://proceedings.neurips.cc/paper/2020/hash/6b493230205f780e1bc26945df7481e5-Abstract.html}
}

@inproceedings{zhang2023speechgpt,
  title     = {{SpeechGPT}: Empowering Large Language Models with Intrinsic Cross-Modal Conversational Abilities},
  author    = {Zhang, Dong and Li, Shimin and Zhang, Xin and Zhan, Jun and Wang, Pengyu and Zhou, Yaqian and Qiu, Xipeng},
  booktitle = {Findings of the Association for Computational Linguistics: EMNLP 2023},
  year      = {2023},
  pages     = {15757--15773},
  publisher = {Association for Computational Linguistics},
  doi       = {10.18653/v1/2023.findings-emnlp.1055},
  url       = {https://aclanthology.org/2023.findings-emnlp.1055/}
}

@inproceedings{tang2024salmonn,
  title     = {{SALMONN}: Towards Generic Hearing Abilities for Large Language Models},
  author    = {Tang, Changli and Yu, Wenyi and Sun, Guangzhi and Chen, Xianzhao and Tan, Tian and Li, Wei and Lu, Lu and Ma, Zejun and Zhang, Chao},
  booktitle = {The Twelfth International Conference on Learning Representations},
  year      = {2024},
  url       = {https://openreview.net/forum?id=t8EF6AAImY}
}

@misc{chu2024qwen2audio,
  title         = {{Qwen2-Audio} Technical Report},
  author        = {Chu, Yunfei and Xu, Jin and Yang, Qian and Wei, Haojie and Wei, Xipin and Guo, Zhifang and Leng, Yichong and Lv, Yuanjun and He, Jinzheng and Lin, Junyang and Zhou, Chang and Zhou, Jingren},
  year          = {2024},
  eprint        = {2407.10759},
  archiveprefix = {arXiv},
  primaryclass  = {cs.CL},
  url           = {https://arxiv.org/abs/2407.10759}
}

@misc{xie2024miniomni,
  title         = {{Mini-Omni}: Language Models Can Hear, Talk While Thinking in Streaming},
  author        = {Xie, Zhifei and Wu, Changqiao},
  year          = {2024},
  eprint        = {2408.16725},
  archiveprefix = {arXiv},
  primaryclass  = {cs.CL},
  url           = {https://arxiv.org/abs/2408.16725}
}

@article{zhong2024memorybank,
  title   = {{MemoryBank}: Enhancing Large Language Models with Long-Term Memory},
  author  = {Zhong, Wanjun and Guo, Lianghong and Gao, Qiqi and Ye, He and Wang, Yanlin},
  journal = {Proceedings of the AAAI Conference on Artificial Intelligence},
  volume  = {38},
  number  = {17},
  year    = {2024},
  pages   = {19724--19731},
  doi     = {10.1609/aaai.v38i17.29946},
  url     = {https://ojs.aaai.org/index.php/AAAI/article/view/29946}
}

@inproceedings{sarthi2024raptor,
  title     = {{RAPTOR}: Recursive Abstractive Processing for Tree-Organized Retrieval},
  author    = {Sarthi, Parth and Abdullah, Salman and Tuli, Aditi and Khanna, Shubh and Goldie, Anna and Manning, Christopher D.},
  booktitle = {The Twelfth International Conference on Learning Representations},
  year      = {2024},
  url       = {https://openreview.net/forum?id=GN921JHCRw}
}

@inproceedings{gutierrez2024hipporag,
  title     = {{HippoRAG}: Neurobiologically Inspired Long-Term Memory for Large Language Models},
  author    = {Jim{\'e}nez Guti{\'e}rrez, Bernal and Shu, Yiheng and Gu, Yu and Yasunaga, Michihiro and Su, Yu},
  booktitle = {Advances in Neural Information Processing Systems},
  volume    = {37},
  year      = {2024},
  doi       = {10.52202/079017-1902},
  url       = {https://proceedings.neurips.cc/paper_files/paper/2024/hash/6ddc001d07ca4f319af96a3024f6dbd1-Abstract.html}
}

@inproceedings{zhang2026memgen,
  title     = {{MemGen}: Weaving Generative Latent Memory for Self-Evolving Agents},
  author    = {Zhang, Guibin and Fu, Muxin and Yan, Shuicheng},
  booktitle = {The Fourteenth International Conference on Learning Representations},
  year      = {2026},
  url       = {https://openreview.net/forum?id=vI56m4Iu4e}
}

@inproceedings{wang2025audiobench,
  title     = {{AudioBench}: A Universal Benchmark for Audio Large Language Models},
  author    = {Wang, Bin and Zou, Xunlong and Lin, Geyu and Sun, Shuo and Liu, Zhuohan and Zhang, Wenyu and Liu, Zhengyuan and Aw, AiTi and Chen, Nancy F.},
  booktitle = {Proceedings of the 2025 Conference of the Nations of the Americas Chapter of the Association for Computational Linguistics: Human Language Technologies (Volume 1: Long Papers)},
  year      = {2025},
  pages     = {4297--4316},
  publisher = {Association for Computational Linguistics},
  doi       = {10.18653/v1/2025.naacl-long.218},
  url       = {https://aclanthology.org/2025.naacl-long.218/}
}

@inproceedings{gosai2026audiomultichallenge,
  title     = {Audio {MultiChallenge}: A Multi-Turn Evaluation of Spoken Dialogue Systems on Natural Human Interaction},
  author    = {Gosai, Advait and Vuong, Tyler and Tyagi, Utkarsh and Li, Steven and You, Wenjia and Bavare, Miheer and U{\c{c}}ar, Arda and Fang, Zhongwang and Jang, Brian and Liu, Bing and He, Yunzhong},
  booktitle = {Proceedings of the 64th Annual Meeting of the Association for Computational Linguistics (Volume 1: Long Papers)},
  year      = {2026},
  pages     = {35740--35770},
  publisher = {Association for Computational Linguistics},
  doi       = {10.18653/v1/2026.acl-long.1654},
  url       = {https://aclanthology.org/2026.acl-long.1654/}
}

@article{moshi,
  title   = {Moshi: A Speech-Text Foundation Model for Real-Time Dialogue},
  author  = {D{\'e}fossez, Alexandre and Mazar{\'e}, Laurent and Orsini, Manu and Royer, Am{\'e}lie and P{\'e}rez, Patrick and J{\'e}gou, Herv{\'e} and Grave, Edouard and Zeghidour, Neil},
  journal = {arXiv preprint arXiv:2410.00037},
  year    = {2024},
  url     = {https://arxiv.org/abs/2410.00037}
}

@inproceedings{freezeomni,
  title     = {Freeze-Omni: A Smart and Low Latency Speech-to-Speech Dialogue Model with Frozen {LLM}},
  author    = {Wang, Xiong and Li, Yangze and Fu, Chaoyou and Zhang, Yike and Shen, Yunhang and Xie, Lei and Li, Ke and Sun, Xing and Ma, Long},
  booktitle = {Proceedings of the 42nd International Conference on Machine Learning},
  series    = {Proceedings of Machine Learning Research},
  volume    = {267},
  pages     = {63345--63354},
  publisher = {PMLR},
  year      = {2025},
  url       = {https://proceedings.mlr.press/v267/wang25aw.html}
}

@misc{minicpmo45,
  title         = {{MiniCPM-o} 4.5: Towards Real-Time Full-Duplex Omni-Modal Interaction},
  author        = {Cui, Junbo and Xu, Bokai and Wang, Chongyi and Yu, Tianyu and Sun, Weiyue and Xu, Yingjing and Wang, Tianran and He, Zhihui and Ma, Wenshuo and Cai, Tianchi and Gui, Jiancheng and Zhang, Luoyuan and Sun, Xian and Huang, Fuwei and Chen, Moye and Lin, Zhuo and Liu, Hanyu and Gui, Qingxin and Han, Qingzhe and Wen, Yuyang and Liu, Huiping and Wang, Rongkang and Zhang, Yaqi and Wei, Hongliang and Chen, Chi and Li, You and Fang, Kechen and Zhou, Jie and Li, Yuxuan and Zeng, Guoyang and Xiao, Chaojun and Lin, Yankai and Han, Xu and Sun, Maosong and Liu, Zhiyuan and Yao, Yuan},
  year          = {2026},
  eprint        = {2604.27393},
  archivePrefix = {arXiv},
  primaryClass  = {cs.CV},
  url           = {https://arxiv.org/abs/2604.27393}
}

@misc{personaplex,
  title         = {{PersonaPlex}: Voice and Role Control for Full Duplex Conversational Speech Models},
  author        = {Roy, Rajarshi and Raiman, Jonathan and Lee, Sang-gil and Ene, Teodor-Dumitru and Kirby, Robert and Kim, Sungwon and Kim, Jaehyeon and Catanzaro, Bryan},
  year          = {2026},
  eprint        = {2602.06053},
  archivePrefix = {arXiv},
  primaryClass  = {eess.AS},
  url           = {https://arxiv.org/abs/2602.06053}
}

@inproceedings{lycheefd,
  title     = {Hierarchical Acoustic-Semantic Modeling: Modality Separation and Semantic Coherence for Full-Duplex {SLM}s},
  author    = {Liu, Zhenyu and Zhang, Xuanyu and Li, Yunxin and Teng, Qixun and Jiang, Shenyuan and Chen, Haolan and Zhao, Mingjun and Meng, Fanbo and Xu, Yu and He, Yancheng and Hu, Baotian and Li, Haizhou and Zhang, Min},
  booktitle = {Proceedings of the 64th Annual Meeting of the Association for Computational Linguistics (Volume 1: Long Papers)},
  pages     = {9264--9280},
  publisher = {Association for Computational Linguistics},
  year      = {2026},
  doi       = {10.18653/v1/2026.acl-long.419},
  url       = {https://aclanthology.org/2026.acl-long.419/}
}

@article{nguyen2023generative,
  title     = {Generative Spoken Dialogue Language Modeling},
  author    = {Nguyen, Tu Anh and Kharitonov, Eugene and Copet, Jade and Adi, Yossi and Hsu, Wei-Ning and Elkahky, Ali and Tomasello, Paden and Algayres, Robin and Sagot, Beno{\^i}t and Mohamed, Abdelrahman and Dupoux, Emmanuel},
  journal   = {Transactions of the Association for Computational Linguistics},
  volume    = {11},
  pages     = {250--266},
  publisher = {MIT Press},
  year      = {2023},
  doi       = {10.1162/tacl_a_00545},
  url       = {https://aclanthology.org/2023.tacl-1.15/}
}

@inproceedings{veluri2024beyond,
  title     = {Beyond Turn-Based Interfaces: Synchronous {LLM}s as Full-Duplex Dialogue Agents},
  author    = {Veluri, Bandhav and Peloquin, Benjamin N. and Yu, Bokai and Gong, Hongyu and Gollakota, Shyamnath},
  booktitle = {Proceedings of the 2024 Conference on Empirical Methods in Natural Language Processing},
  pages     = {21390--21402},
  publisher = {Association for Computational Linguistics},
  year      = {2024},
  doi       = {10.18653/v1/2024.emnlp-main.1192},
  url       = {https://aclanthology.org/2024.emnlp-main.1192/}
}

@inproceedings{wang2025fullduplex,
  title     = {{NTPP}: Generative Speech Language Modeling for Dual-Channel Spoken Dialogue via Next-Token-Pair Prediction},
  author    = {Wang, Qichao and Meng, Ziqiao and Cui, Wenqian and Zhang, Yifei and Wu, Pengcheng and Wu, Bingzhe and King, Irwin and Chen, Liang and Zhao, Peilin},
  booktitle = {Proceedings of the 42nd International Conference on Machine Learning},
  series    = {Proceedings of Machine Learning Research},
  volume    = {267},
  pages     = {63912--63928},
  publisher = {PMLR},
  year      = {2025},
  url       = {https://proceedings.mlr.press/v267/wang25by.html}
}
